\documentclass[lettersize,journal]{IEEEtran}

\usepackage{array}
\usepackage[caption=false,font=normalsize,labelfont=sf,textfont=sf]{subfig}
\usepackage{textcomp}
\usepackage{stfloats}
\usepackage{verbatim}
\usepackage{graphicx}
\usepackage{balance}
\usepackage{color}
\usepackage{algorithmicx}
\usepackage{algorithm}
\usepackage[colorlinks,urlcolor=blue,linkcolor=blue,citecolor=blue]{hyperref}
\usepackage[T1]{fontenc}    % use 8-bit T1 fonts
\usepackage{amsmath,amsfonts}
\usepackage{array}
\usepackage{booktabs}
\usepackage{bigstrut}
\usepackage{multirow}
\usepackage{textcomp}
\usepackage{url}
\usepackage{cite}
\usepackage{color}
\usepackage{bbding}
\usepackage{ifthen}
\usepackage{algpseudocode}
\usepackage{multirow}
\usepackage{ragged2e}

\begin{document}

\title{B2Opt: Learning to Optimize Black-box Optimization with Little Budget}

\author{Xiaobin Li, \IEEEmembership{Member, IEEE}, Kai Wu, \IEEEmembership{Member, IEEE}, and Xiaoyu Zhang, Handing Wang, \IEEEmembership{Member, IEEE}, Jing Liu, \IEEEmembership{Senior Member, IEEE}
%\thanks{This work was supported by the Key Project of Science and Technology Innovation 2030 supported by the Ministry of Science and Technology of China under Grant 2018AAA01013.}
\thanks{This work was supported in part by the National Natural Science Foundation of China under Grant 62206205, in part by the Fundamental Research Funds for the Central Universities under Grant XJS211905, in part by the Guangdong High-level Innovation Research Institution Project under Grant 2021B0909050008, and in part by the Guangzhou Key Research and Development Program under Grant 202206030003. (\textit{Corresponding author: Jing Liu.})}
\thanks{Xiaobin Li, Kai Wu, and handing Wang are with School of Artificial Intelligence, Xidian University, Xi'an 710071, China (e-mail: kwu@xidian.edu.cn; 22171214784@stu.xidian.edu.cn; hdwang@xidian.edu.cn).}
\thanks{Xiaoyu Zhang is with School of Cyber Engineering, Xidian University, Xi'an 710071, China (e-mail: xiaoyuzhang@xidian.edu.cn).}
\thanks{Jing Liu is with Guangzhou Institute of Technology, Xidian University, Guangzhou 510555, China (e-mail: neouma@mail.xidian.edu.cn).}}

% The paper headers
\markboth{Journal of \LaTeX\ Class Files,~Vol.~14, No.~8, August~2021}%
{Shell \MakeLowercase{\textit{et al.}}:B2Opt: Learning to Optimize Black-box Optimization with Little Budget}

% \IEEEpubid{0000--0000/00\$00.00~\copyright~2021 IEEE}
% Remember, if you use this you must call \IEEEpubidadjcol in the second
% column for its text to clear the IEEEpubid mark.

% todo:
% 1.    修改Fig. 1
% 2.     Tabel. 3 验证标记为红色的实验结果，补充实验结果
% 3.    实验已经修改，需要把对应的实验描述也进行修改
% 4.    在附录部分补充未添加的CEC函数实验结果表格

\maketitle
\begin{abstract}
The core challenge of high-dimensional and expensive black-box optimization (BBO) is how to obtain better performance faster with little function evaluation cost. The essence of the problem is how to design an efficient optimization strategy tailored to the target task. This paper designs a powerful optimization framework to automatically learn the optimization strategies from the target or cheap surrogate task without human intervention. However, current methods are weak for this due to poor representation of optimization strategy. To achieve this, 1) drawing on the mechanism of genetic algorithm, we propose a deep neural network framework called B2Opt, which has a stronger representation of optimization strategies based on survival of the fittest; 2) B2Opt can utilize the cheap surrogate functions of the target task to guide the design of the efficient optimization strategies. Compared to the state-of-the-art BBO baselines, B2Opt can achieve multiple orders of magnitude performance improvement with less function evaluation cost. We validate our proposal on high-dimensional synthetic functions and two real-world applications. We also find that deep B2Opt performs better than shallow ones.
% so what: 我们为可学习的优化策略提供了新的表征方式。
\end{abstract}

\begin{IEEEkeywords}
 Learning to Optimize, Black-box Optimization, Transformer, Learned Optimizer, Genetic Algorithm.
\end{IEEEkeywords}

\section{Introduction}
\IEEEPARstart{M}{any} tasks, such as neural architecture search \cite{elsken2019neural} and hyperparameter optimization \cite{hutter2019automated,golovin2017google}, can be abstracted as black-box optimization (BBO) \cite{li2022evolutionary}, which means that although we can evaluate $f(x)$ for any $x\in X$, we have no access to any other information about $f$, such as the Hessian and gradients. Moreover, it is expensive to evaluate $f(x)$ in most cases. Various hand-designed algorithms, such as genetic algorithms (GAs) \cite{mitchell1998introduction,NEURIPS2018_85fc37b1,zhang2007moea}, Bayesian optimization \cite{Snoek2012Practical,mutny2018efficient,cheng2017high,kandasamy2015high,balandat2020botorch}, and evolution strategy (ES) \cite{wierstra2014natural,hansen2001completely,Auger2005restart,salimans2017evolution}, have been designed to solve BBO. Their purpose is to design optimization strategies that allow them to continuously sample better solutions.

Recently, the learning to optimize (L2O) framework \cite{andrychowicz2016learning,chen2021learning} gives a new insight into optimization. They leverage the recurrent neural network (RNN), long short-term memory architecture (LSTM) \cite{Chen2020Training,chen2017learning,li2016learning,wichrowska2017learned,bello2017neural}, multilayer perceptron (MLP) \cite{metz2019understanding}, or Transformer \cite{TransformerL2O} as the optimizer to develop optimization methods, aiming at reducing the laborious iterations of hand engineering \cite{vicol2021unbiased,flennerhag2021bootstrapped,li2016learning,sun2018optimize}. The core of L2O is constructing a solid mapping from the initial solutions to the optimal solution. Moreover, several efforts \cite{cao2019learning,chen2017learning,lange2022discovering} have coped with the black-box problems, their effectiveness may be hindered by the limited representation capabilities of optimization strategy and a large number of evaluations on expensive black-box functions. As such, they often deal with low-dimensional problems.

%GA中由交叉、变异和选择算子的顺序排列组成了处理黑盒优化的优化策略。但是它们固定的参数引起其抽样策略低效。本文通过来参数化这些模块来获得强优化策略表征。

The optimization strategy in GA consists of a sequential arrangement of crossover, mutation, and selection operators to handle black-box optimization. But their fixed parameters cause inefficient sampling strategies. In this paper, we obtain a strong optimization strategy representation by coming to parameterize these modules, termed as B2Opt.

We design three modules to realize the mapping from the random solution to the optimal solution. To generate potential individuals to approach the optimal solution, we first develop a self-attention (SA)-based crossover module (SAC) to create potential individuals to approach the optimal solution. Then the output of this module is input into the proposed feed-forward network (FFN)-based mutation module (FM) to escape the optimal local solution. Moreover, the residual and selection module (RSSM) is designed to survive the fittest individuals. RSSM is a pairwise comparison between the output of SAC, FM, and the input population regarding their fitness. We develop a B2Opt Block (OB) consisting of SAC, FM, and RSSM. Finally, we simulate the iteration rule of GAs by stacking OBs to represent strong optimization strategies.

In order to adapt the optimization strategy represented by B2Opt to the target task, we update the parameters of B2Opt based on the information of target task. Moreover, to reduce the number of evaluations to an expensive black-box function, we establish a cheap surrogate function set to train B2Opt. We construct a set of cheap differentiable functions with similar properties to the targeted BBO problems. This training set contains the pair of the initial population and the designed surrogate function. Thus, we can use gradient-based methods, such as stochastic gradient descent and Adam \cite{kingma2014adam}, to train B2Opt. In this way, we don't need to query expensive functions during training.

We test B2Opt on six standard functions with high dimension, the neural network training problem, and the planar mechanic arm problem \cite{wang2022solving}. The experimental results demonstrate the top rank of B2Opt and the strong representation compared with five state-of-the-art (SOTA) EA baselines, two SOTA Bayesian baselines, and three SOTA L2O methods. The highlights of this paper are shown as follows:
\begin{itemize}
    \item Compared with the state-of-the-art BBO methods, B2Opt achieves multiple orders of magnitude performance improvement with fewer function evaluations even if the training dataset is a low-fidelity set of the target black-box function.
    \item We design the B2Opt framework to realize the automated GA, which can better represent optimization strategies. It is easier to map random solutions to optimal solutions with fewer evaluation times during optimization.
    \item We design a new training strategy to reduce the evaluation cost of expensive objective functions during training. It uses cheap surrogate functions for expensive target tasks instead of directly evaluating the target black-box function.
\end{itemize}

The rest of this paper is organized as follows. Section II summarizes the related work about learnable optimization strategy. Section III formalizes the studied problem and introduces the background of GAs and Transformer. Section IV presents the content of B2Opt. Section V conducts extensive experiments to evaluate the effectiveness of B2Opt. Finally, Section VI concludes the whole paper.

\section{Related Work}

We mainly focus on a learnable optimization strategy for BBO, which is divided into two parts.

\subsection{Meta-Learn Whole BBO Algorithm}
Our proposal belongs to this type. Chen \emph{et al.} \cite{chen2017learning} first explored meta-learning entire algorithms for low-dimensional BBO. Then, TV \emph{et al.} \cite{tv2020meta} proposed RNN-Opt, which learned recurrent neural network (RNN)-based optimizers for optimizing real parameter single-objective continuous functions under limited budget constraints. Moreover, they train RNN-Opt with knowledge of task optima. Swarm-inspired meta-optimizer \cite{cao2019learning} learns in the algorithmic space of both point-based and population-based optimization algorithms. Gomes \emph{et al.} \cite{gomes2021meta} employed meta-learning to infer population-based black-box optimizers that can automatically adapt to specific classes of tasks. These methods parametrize the optimization algorithm by an RNN processing the raw solution vectors and their associated fitness. 

\subsection{Meta-Learn Part BBO Algorithm}
This type only learns some parameters of the algorithm, not the overall algorithm.
Shala \emph{et al.} \cite{shala2020learning} meta-learned a policy that configured the mutation step-size parameters of CMA-ES \cite{hansen2001completely}. LES \cite{lange2022discovering} designed a self-attention-based search strategy to discover effective update rules for evolution strategies via meta-learning. Subsequent
work LGA \cite{lange2023discovering} employed the above framework to discover the update rules of Gaussian genetic algorithms via Open-ES \cite{salimans2017evolution}.

The limitations of these methods are shown as follows: 1) to train the meta-optimizer, a large number of expensive black-box functions need to be requested, which is very unrealistic; 2) the established loss function for training the meta-optimizer is challenging to optimize, resulting in a poor representation of the optimization strategy. Thus, we propose B2Opt overcome the above limitations.

\section{Preliminaries}
%%%%%%%%%%%%背景知识%%%%%%%%%%

%%%%%在这里写清楚transformer,EA和meta-learning.
\subsection{Genetic Algorithms}
The crossover, mutation, and selection operators form the basic framework of GAs. GA starts with a randomly generated initial population. Then, genetic operations such as crossover and mutation will be carried out. After the fitness evaluation of all individuals in the population, a selection operation is performed to identify fitter individuals to undergo reproduction to generate offspring. Such an evolutionary process will be repeated until specific predefined stopping criteria are satisfied.

\paragraph{Crossover} The crossover operator generates a new individual $\hat{X_i}$ by Eq. (\ref{eq:2}), and $cr$ is the probability of the crossover operator.
\begin{equation}\label{eq:2}
\hat{X}_{i,k}^c = \left\{
\begin{matrix}
X_{j,k}& rand(0,1) < cr\\
X_{i,k} & otherwise
\end{matrix} \right.
\end{equation}
where $k \in [1, \cdots, d]$. This operator is commonly conducted on $n$ individuals. After an expression expansion, we re-formulate Eq. (\ref{eq:2}) as $\sum_{i=1}^n X_i W_i^c$ \cite{zhang2021analogous}. $W_i^c$ is the diagonal matrix. If $W_i^c$ is full of zeros, the $i$th individual has no contribution.

\paragraph{Mutation}
The mutation operator brings random changes into the population. Specifically, an individual
$X_i$ in the population goes through the mutation operator to form the new individual $\hat{X}_i$, formulated as follows:
\begin{equation}\label{eq:3}
\hat{X}_{i,k}^m = \left\{
\begin{matrix}
rand(l_k, u_k)& rand(0,1) < mr\\
\hat{X}_{i,k}^c & otherwise
\end{matrix} \right.
\end{equation}
where $mr$ is the probability of mutation operator and $k \in [1, \cdots, d]$. Similarly, Equation (\ref{eq:3}) can be re-formulated as $X_i W_i^m$, where $W_i^m$ is the diagonal matrix.

\paragraph{Selection} 
We introduce the binary tournament mating selection operator in Eq. (\ref{eq:4}). The selection operator survives individuals of higher quality for the next generation until the number of individuals is chosen.
\begin{equation}\label{eq:4}
p_i = \left\{
\begin{matrix}
1 & f(X_i) < f(X_k)\\
0 & f(X_i) > f(X_k)
\end{matrix} \right.
, \ \ (X_i,X_k) \in X,
\end{equation}
where $p_i$ reflects the probability that $X_i$ is selected for the next generation, and $(X_i,X_k)$ in Eq. \ref{eq:4} are randomly selected from the population $X \cup \hat{X}^m$.

\begin{figure*}
%\vskip 0.2in
\centerline{\includegraphics[width=0.95\linewidth]{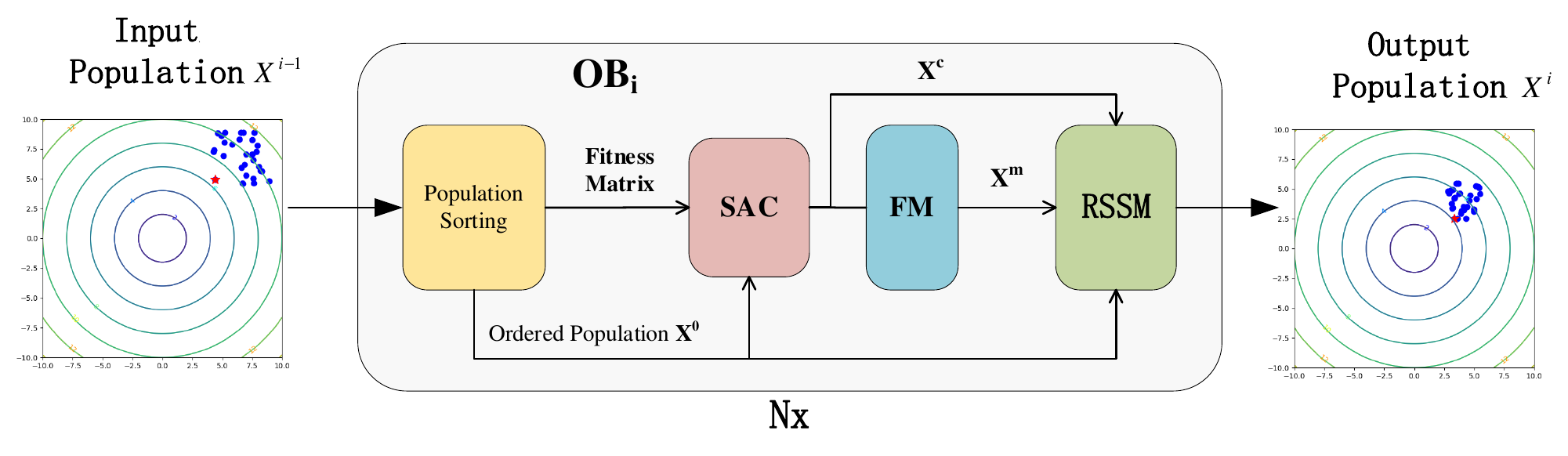}}
\caption{Overall architecture of B2Opt and OB. $Nx$ stands for B2Opt is composed of $Nx$  stacked OBs. These OBs can be set to share weights with each other or not share weights with each other.}
%\vskip -0.2in
\label{fig:structure}
\end{figure*}

\subsection{Transformer}

We mainly introduce the core part of Vision Transformer \cite{dosovitskiy2021image}, such as the multi-head self-attention layer (MSA), feed-forward network (FFN), layer normalization (LN), and residual connection (RC). 

\subsubsection{MSA}
MSA fuses several SA operations to handle the queries ($Q$), keys ($K$), and values ($V$)
that jointly attend to information from different representation subspaces. MSA is formulated as follows: $MultiHead(Q, K, V) = Concat(H_1, H_2, \cdots H_h)W^O$. where $Concat$ means concatenation operation. The head feature $H_i$ can be formulated as:
\begin{align*}\label{eq:5}
H_i &= SA(QW_i^Q, KW_i^K, VW_i^V)\\
      &= Softmax\left(QW_i^Q(KW_i^K)^T/sqrt(d_k)\right)VW_i^V = AVW_i^V
\end{align*}
where $W_i^Q \in R^{d_m \times d_q}$, $W_i^K \in R^{d_m \times d_k}$, and $W_i^V \in R^{d_m \times d_v}$ are parameter matrices for queries, keys, and values, respectively; $W^O \in R^{hd_v \times d_m}$ maps each head feature $H_i$ to the output. Moreover, $d_m$ is the input dimension, while $d_q$, $d_k$, and $d_v$ are hidden dimensions of the corresponding projection subspace; $h$ is the head number. $A \in R^{l \times l}$ is the attention matrix of $h$th head, $l$ is the sequence length.

\subsubsection{FFN}
FFN employs two cascaded linear transformations with a ReLU activation to handle $X$, which is shown as: 
$FFN(X) = max(0,XW_1 + b_1)W_2 + b_2$, 
where $W_1$ and $W_2$ are weights of two linear layers, and $b_1$ and $b_2$ are corresponding biases.
\subsubsection{LN}
LN is applied before each layer of MSA and FFN, and the output of LN is calculated by $X + [MSA|FFN](LN(X))$.

\section{B2Opt}

\subsection{Problem Definition}
A black-box optimization problem can be transformed as a minimization problem, as shown in Eq. (\ref{eq:1}), and constraints may exist for corresponding solutions:
\begin{equation}\label{eq:1}
\min_x \ f(x), s.t. \ x_i \in [l_i,u_i]%, \forall x_i \in s,
\end{equation}
where $x = (x_1, x_2, \cdots, x_d)$ represents the solution of optimization problem $f$, the lower and upper bounds $l = (l_1, l_2,\cdots, l_d)$ and $u = (u_1, u_2, \cdots, u_d)$, and $d$ is the dimension of $x$. Suppose $n$ individuals of one population be $X_1=(X_{1,1}, X_{1,2},\cdots, X_{1,d}), \cdots, X_n=(X_{n,1}, X_{n,2},\cdots, X_{n,d})$, then B2Opt are required to find the population near the optimal solution. Note that we only have a very small number of function evaluations to achieve.
%We suppose that $X^0$ is the initial population. and $X^t$ is the output population.

\subsection{Self-Attention Crossover Module}

Inspired by the crossover operator in GAs, we propose an SA-based module to generate potential solutions by maximizing information interaction among individuals in a population. %The crossover operator generates a new individual can be written as $\sum_{i=1}^n X_i W_i^c$ \cite{zhang2021analogous,zhang2022eatformer}. $W_i^c$ is the diagonal matrix, which is fixed in traditional GAs. If $W_i^c$ is full of zeros, the $i$th individual has no contribution.
Suppose a population $X$ is arranged in a non-descending order of fitness, and $F \in R^{n \times 1}$ be the fitness matrix of $X$. Then, this module can be represented as follows:
\begin{equation}\label{eq:6}
X^c = SAC(X, F)
\end{equation}
where $X^c$ is the output population of the SAC module.

Since the object processed by B2Opt is the population with several individuals, and the order of individuals does not affect the population distribution, SA does not require position coding. Standard SA projects the input sequence $X$ into a $d$-dimensional space via the queries ($Q$), keys ($K$), and values ($V$). These three mappings enable the SA module to better capture the characteristics of the problems encountered during training. In other words, these three mappings strengthen the ability of SA to focus on specific issues but do not necessarily make SA have good transferability between different problems. Therefore, we consider removing these three mappings for enhanced transferability and $X^c = AX$. $A \in R^{n \times n}$ is a self-attention matrix that can be learned to maximize inter-individual information interaction based on individual ranking information. Since $A$ uses individual ranking information instead of fitness value information, $A$ is invariant to the problems, which is beneficial to improve the generalization performance of the model. This is why the population needs to be sorted in non-descending order.

However, designing the SAC module based solely on population ranking information is a coarse-grained approach. Because this way only considers the location information of individuals in the population but does not consider the fitness relationship between individuals. Therefore, we further introduce fitness information to assist in designing the SAC module:
\begin{equation}\label{eq:sac}
A^F = SA(F) = Softmax\left(FW^Q(FW^K)^T/sqrt(d_k)\right)
\end{equation}
Thus, $X^c = AX + A^FX$. To better balance the roles of $A$ and $A^F$, we introduce two learnable weights $W_1^c \in R^{n \times 1}$ and $W_2^c \in R^{n \times 1}$. Therefore, the final SAC module is shown as follows: 
\begin{equation}\label{eq:7}
X^c = tile(W_1^c) \odot (AX) + tile(W_2^c) \odot (A^FX)
\end{equation}
where $\odot$ represents Hadamard product and the \emph{tile} copy function extends the vector to a matrix.

\subsection{FFN-based Mutation Module}
 
%the mutation operator in GAs brings random changes into the population. Specifically, an individual $X_i$ in the population goes through the mutation operator to form the new individual $\hat{X}_i$, formulated as $\hat{X}_i = X_i W_i^m$. $W_i^m$ is the diagonal matrix.
In Transformer, each patch embedding carries on directional feature transformation through the FFN module. We take one linear layer as an example: $X = XW^F$, 
where $W^F$ is the weight of the linear layer, and it is applied to each embedding separately and identically. The similar formula format of mutation operator and FFN inspires us to design a learnable mutation module FM based on FFN with $ReLU$ activation function:
\begin{equation}\label{eq:8}
%\begin{aligned}
X^m = FM(X^c) = (ReLU(X^cW_1^F+b_1))W_2^F+b_2
%\end{aligned}
\end{equation}
where $X^m$ is the population after the mutation of $X^c$. $W_2^F$ and $W_1^F$ represent the weight of the second layer of FFN and the weight of the first layer of FFN, respectively. $b_2$ and $b_1$ represent the bias of the second layer and the first layer of FFN, respectively. FM is designed to escape the local optimum.
%Moreover, in $b_2$ or $b_1$, the elements in $\{i,k\}$ is equal to $\{j,k\}, \left(k\in\left[0,d\right)\right)$.

\subsection{Selection Module}
The residual connection in the transformer can be analogized to the selection operation in GA \cite{zhang2021analogous}. 
%We combine the residual structure and selection module (SM) \cite{anonymous2023decn} to design a learnable selection module RSSM.
The RSSM generates the offspring population according to the following equation:
\begin{equation}\label{eq:9}
\begin{aligned}
\hat{X} &= RSSM(X, X^c, X^m) \\
&= Sort(SM(X, tile(W_1^s) \odot X \\
&+tile(W_2^s) \odot X^c+tile(W_3^s) \odot X^m))
\end{aligned}
\end{equation}
where $\hat{X}$ is the fittest population for the next generation; the learnable weights $W_1^s \in R^{n \times 1}$, $W_2^s \in R^{n \times 1}$, and $W_3^s \in R^{n \times 1}$ are the weights for $X$, $X^c$, and $X^m$, respectively. %$\bullet$ indicates the pairwise multiplication between inputs.
$Sort(X)$ represents that $X$ is sorted in non-descending order of fitness. We use quicksort to sort the population based on function fitness. These three learnable weight matrices realize the weighted summation of residual connections, thereby simulating a learnable selection strategy. Meanwhile, introducing residual structure enhances the model's representation ability, enabling B2Opt to form a deep architecture.

SM updates individuals based on a pairwise comparison between the offspring and input population regarding their fitness. Suppose that $X$ and $X^{'}$ are the input populations of SM. We compare the quality of individuals from $X$ and $X^{'}$ pairwise based on fitness. A binary mask matrix indicating the selected individual can be obtained based on the indicator function $l_{x>0}(x)$, where $l_{x>0}(x)=1$ if $x>0$ and $l_{x>0}(x)=0$ if $x<0$. SM forms a new population $\hat{X}$ by employing Eq. (\ref{eq:12}).
\begin{equation}\label{eq:12}
\begin{split}
\hat{X} &= tile(l_{x>0}(M_{F'}-M_F)) \odot X \\
&+ tile(1-l_{x>0}(M_{F'}-M_F)) \odot X^{'}
\end{split}
\end{equation}
where the \emph{tile} copy function extends the indication vector to a matrix, $M_F (M_{F'})$ denotes the fitness matrix of $X (X^{'})$.

\subsection{Structure of B2Opt}

B2Opt comprises basic $t$ B2Opt blocks (OBs), and parameters can be shared among these $t$ OBs or not. The overall architecture of B2Opt and OB is shown in Figure \ref{fig:structure}. Each OB consists of SAC, FM, and RSSM in sequential order. $X^0 \in R^{n\times d}$ represents the initial population input into B2Opt, which must be sorted in non-descending fitness order. In Eq. \ref{eq:total}, $X^{i-1}$ is fed into $OB_t$ to get $X^i$, where $i \in [1,t]$. B2Opt realizes the mapping from the random initial population to the target population by stacking $t$ OBs.  
\begin{gather} \label{eq:total}
X^i = OB(X^{i-1}); \ \ \ X^c  = SAC(X^{i-1}, F); \\
X^{m}  = FM(X^{c});\ \ \ X^i = RSSM(X^{i-1}, X^c, X^m) \nonumber
\end{gather}

\subsection{Training of B2Opt}

\subsubsection{Goal}
We introduce the parameters $\theta$ of B2Opt, which need to be optimized. Here, we set $\theta = \{W_1^c, W_2^c, A, W_1^F, W_2^F, b_1, b_2, W_1^s, W_2^s, W_3^s\}$.

\subsubsection{Training Dataset}

Before introducing the details of the training dataset, fidelity \cite{kandasamy2016gaussian} is defined as follows:
Suppose the differentiable surrogate functions $f_1, f_2, \cdots, f_m$ are the continuous exact approximations of the black-box function $f$. We call these approximations fidelity, which satisfies the following conditions:
1) $f_1, \cdots, f_i, \cdots, f_m$ approximate $f$. $||f-f_i||_{\infty} \leq \zeta_m$, where the fidelity bound $\zeta_1>\zeta_2> \cdots \zeta_m$.
2) Estimating approximation $f_i$ is cheaper than estimating $f$. Suppose the query cost at fidelity is $\lambda_i$, and $\lambda_1<\lambda_2< \cdots \lambda_m$.

Training data is a crucial factor beyond the objective functions. This paper establishes the training set by constructing a set of differentiable functions related to the optimization objective.
This training dataset only contains $(X_0, f_i(x|\omega))$, the initial population and objective function, respectively. The variance of $\omega$ causes the shift in landscapes. The training dataset is designed as follows: 
\begin{enumerate}
    \item Randomly initialize the input population $X_0$;
    \item Randomly produce a shifted objective function $f_i(x|\omega)$ by adjusting the parameter $\omega$;
    \item Evaluate $X_0$ by $f_i(x|\omega)$;
    \item Repeat Steps 1)-3) to generate the corresponding dataset. We show the designed training and testing datasets as follows:
\end{enumerate}
   
\begin{equation}\label{eq:19}
F^{train} = \{ f_1(x|\omega_{1,i}^{train}),\cdots, f_m(x|\omega_{m,i}^{train})\}
\end{equation}
where $\omega_{m,i}^{train}$ represents the $i$th different values of $\omega$ in the $m$th function $f_m$.

\subsubsection{Loss Function}
B2Opt attempts to search for individuals with high quality based on the available information. 
The loss function tells how to obtain the parameters of B2Opt to generate individuals closer to the optimal solution by maximizing the difference between the initial population and the output population of B2Opt. The following loss function $l_i(X^0, f(x|\omega), \theta)$ is employed, %\cite{anonymous2023decn},
\begin{equation}\label{eq:14}
l_i = \frac{ \frac{1}{|X^0|} {\sum \limits_{x\in X^0} {f_i(x|\omega)}} -\frac{1}{|E_{\theta}(X^0)|} {\sum \limits_{x \in E_{\theta}(X^0)}} f_i(x|\omega)}{\left|{\frac{1}{|X^0|} {\sum \limits_{x\in X^0} f_i(x|\omega)}}\right|}
\end{equation}
where $\theta$ denotes parameters of B2Opt ($E$). 
Equation (\ref{eq:14}) calculates the average fitness difference between the input and output, further normalized within $[0, 1]$. To encourage B2Opt to explore the fitness landscape, for example, the constructed Bayesian posterior distribution over the global optimum \cite{cao2020bayesian} can be added to Eq. (\ref{eq:14}). We have three ways to train B2Opt:
\begin{enumerate}
    \item When we can construct a differentiable proxy function of the objective function, we use the gradient information of Eq. (\ref{eq:14}) to train B2Opt.
    \item When it is difficult to construct differentiable surrogate functions of the objective function, we can use REINFORCE \cite{williams1992simple} to estimate these derivatives.
    \item We can use ES \cite{vicol2021unbiased} to train B2Opt. Here, we focus on introducing the training of B2Opt through the first method.
\end{enumerate}

\subsubsection{Training B2Opt}

We then train B2Opt under a supervised mode. Since the gradient is unnecessary during the test process, B2Opt can solve BBO problems. To prepare B2Opt to learn a balanced performance upon different optimization problems, we design a loss function formulated as follows:
\begin{equation}\label{eq:16}
\underset{\theta}{\arg\min} \ \ l_{\Omega} = -\frac{1}{K} \sum_{X^0 \in \Omega} {l_i(X^0,f_i(x|\omega_i^{train}), \theta)}
\end{equation}
We employ Adam \cite{kingma2014adam} method with a minibatch $\Omega$ to train B2Opt upon training dataset. 

\subsubsection{Detailed Training Process} The goal of the training algorithm is to search for parameters ${\theta}^*$ of the B2Opt. Before training starts, B2Opt is randomly initialized to get initial parameters $\theta$. Then the algorithm will perform the following three steps in a loop until the training termination condition is satisfied: 
\begin{enumerate}
    \item \textbf{Step 1}, randomly initialize a minibatch $\Omega$ comprised of $K$ populations $X^0$;
    \item \textbf{Step 2}, for each $f_i \in {F}^{train}$, given training data $(X^0, f_i)$, update $\theta$ by minimizing the $l_{\Omega}$;
    \item \textbf{Step 3}, given $X^0$, update $\theta$ by minimizing $-{1}/{m} \sum_i l_{\Omega}$, where $m$ is the number of functions in ${F}^{train}$. After completing the training process, the algorithm will output ${\theta}^*$.
\end{enumerate}

\section{Experiments}
\subsection{Experimental Setup}
\subsubsection{Datasets}

%testing functions
\begin{table*}[htb]
\caption{Training and testing functions. F1-F3 are training functions and F4-F9 are testing functions. Here, $d=\{10, 100\}$. $z_i=x_i-b_i$.}
\label{table:test}
\begin{center}
%{lp{9.5cm}c}
\begin{tabular}{ccc}
\toprule
ID & Functions & Range \\
\midrule
F1    & $\sum_i {|w_i sin(x_i-b_i)|}$ & $x \in [-10,10],b\in [-10,10]$ \\
F2    & $\sum_i {|x_i-b_i|}$ & $x \in [-10,10], b\in [-10,10]$ \\
F3    & $\sum_i {|(x_i-b_i)+(x_{i+1}-b_{i+1})|} + \sum_i {|x_i-b_i|}$ & $x \in [-10,10],b\in [-10,10]$ \\
\midrule
F4(Sphere)    & $\sum_i z_i^2$ & $x \in [-100,100],b\in [-50,50]$ \\
F5    & $\{|z_i|, 1 \le i \le D\}$ & $x \in [-100,100],b\in [-50,50]$ \\
F6(Rosenbrock)    & $\sum \limits_{i=1}^{D-1} (100(z_i^2-z_{i+1})^2+{(z_i-1)}^2)$ & $x \in [-100,100],b\in [-50,50]$ \\
F7(Rastrigin)    & $\sum \limits_{i=1}^{D} (z_i^2-10\cos(2\pi z_i)+10)$ & $x \in [-5,5],b\in [-2.5,2.5]$ \\
F8(Griewank)    & $\sum \limits_{i=1}^{D} {\frac{z_i^2}{4000}}- \prod_{i=1}^D \cos(\frac{z_i}{\sqrt{i}})+1$ & $x \in [-600,600], b\in [-300,300]$ \\
F9(Ackley)    & $-20\exp(-0.2\sqrt{\frac{1}{D} \sum_{i=1}^D z_i^2})-\exp(\frac{1}{D} \sum_{i=1}^D \cos(2\pi z_i))+20+\exp(1)$ & $x \in [-32,32],b\in [-16,16]$ \\
\bottomrule
\end{tabular}
\end{center}
\end{table*}

\paragraph{Synthetic Functions}
This paper first employs nine commonly used functions to show the effectiveness of the proposed B2Opt. The characteristics of these nine functions are shown in Table \ref{table:test}. Here, B2Opt is trained on $F^{train}$ generated based on F1-F3, and the target functions are F4-F9. B2Opt is trained on the set of tasks with low fidelity for the target function. We test the generalization performance of B2Opt through this case. Moreover, B2Opt is trained on target functions with different function biases. We then test it on target function with biases unseen in the training stage; we refer it to B2Opt-V2.

\begin{figure}[!t]
%\vskip 0.2in
\begin{center}
\centerline{\includegraphics[width=0.9\linewidth]{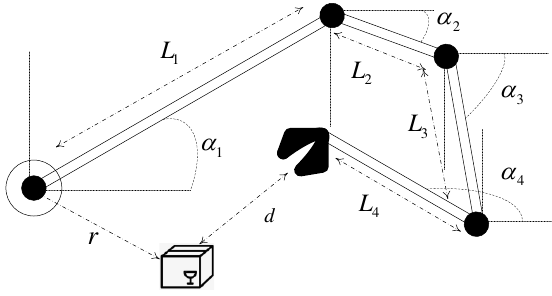}}
\caption{\centering{Planar Mechanical Arm.}}
\label{fig:pma}
\end{center}
%\vskip -0.2in
\end{figure}

\paragraph{Planner Mechanic Arm}
We further evaluate the performance of the proposed scheme on the planner mechanic arm problem \cite{Cully2015RobotsTC,Vassiliades2018UsingCV,Vassiliades2018DiscoveringTE,Mouret2020QualityDF}, a robotic control problem that has been widely used to evaluate the performance of BBO algorithms. It is shown in Fig. \ref{fig:pma}. The optimization goal of this problem is to search for a set of lengths $L=(L_1, L_2, \cdots, L_n)$ and a set of angles $\alpha=(\alpha_1, \alpha_2, \cdots, \alpha_n)$ so that the distance $f(L, \alpha, p)$ from the top of the mechanic arm to the target position $p$ is the smallest, where $n$ represents the number of segments of the mechanic arm, and $L_i \in (l_i, u_i)$ and $\alpha_i \in (-\Pi, \Pi)$ represent the length and angle of the $i$th mechanic arm, respectively. $r$ represents the distance from the target point to the origin of the mechanic arm. Typically, $f=\sqrt{\left(\sum_{i=1}^n \cos(\alpha_i)L_i-p_x\right)^2 + \left(\sum_{i=1}^n \sin(\alpha_i)L_i-p_y\right)^2}$, where $p_x$ and $p_y$ represent the $x$-coordinate and $y$-coordinate of the target point, respectively. 

Here, $n=100$, $l_i=0$ and $u_i=10$. We design two groups of experiments:\\
\textit{1) Simple Case}. We fixed the length of each mechanic arm as $l_i=10$ and only searched for the optimal $\alpha$.\\
\textit{2) Complex Case}. We need to search for $L$ and $\alpha$ simultaneously. We randomly selected 600 target points within the range of $r \leq 1000$ to form a set $S$, where $r$ represents the distance from the target point to the origin of the mechanic arm, as shown in Fig. \ref{fig:pma}. During the training process of B2Opt, a sample point set $s$ is re-extracted from $S$ for training every $T$ training cycle. In the testing process, we extracted 128 target points ($S^{test}$) in the range of $r \leq 100$, $r \leq 300$, and $r \leq 1000$, respectively, for testing. The purpose of testing in three different regions is to explore the generalization performance of B2Opt further. We evaluate the generalization ability of the algorithm by $\left(\sum_s^{S^{test}} f(L, \alpha, s)\right) / |S^{test}|$.

\begin{table}[htbp]
  \centering
  \small
  \caption{The small-scale classification neural network for MNIST.}
  \resizebox{0.5\textwidth}{!}{
    \begin{tabular}{ccccc}
    \toprule
    Layer ID & Layer Type & Padding & Stride & Kernel Size \\
    \midrule
    1     & depth-wise convolution & \XSolidBrush & 1 & 1$\times$5$\times$5 \bigstrut\\
    2     & point-wise convolution & \XSolidBrush &1 & 3$\times$1$\times$1$\times$1 \bigstrut\\
    3     & ReLu  & \XSolidBrush  & \XSolidBrush  & \XSolidBrush \bigstrut\\
    4     & max pooling & \XSolidBrush & 2 & 2$\times$2 \bigstrut\\
    5     & depth-wise convolution & \XSolidBrush & 1 & 3$\times$5$\times$5 \bigstrut\\
    6     & point-wise convolution & \XSolidBrush & 1 & 16$\times$3$\times$1$\times$1 \bigstrut\\
    7     & ReLu  & \XSolidBrush  & \XSolidBrush  & \XSolidBrush \bigstrut\\
    8     & max pooling & \XSolidBrush & 2 & 2$\times$2 \bigstrut\\
    9     & depth-wise convolution & \XSolidBrush & 1 & 16$\times$4$\times$4 \bigstrut\\
    10    & point-wise convolution & \XSolidBrush & 1 & 10$\times$16$\times$1$\times$1 \bigstrut\\
    11    & softmax & \XSolidBrush  & \XSolidBrush  & \XSolidBrush \bigstrut\\
    \bottomrule
    \end{tabular}}
  \label{tab:mnist net}
\end{table}

%标红部分，如果审稿人问：“为什么替换为cnn；cnn和SAC模块有什么关系；不替换的效果怎么样”，该怎么回答？是否有必要
%在这个问题做更多的解释？有没有可能，我们说使用CNN作为SAC模块是我们的变体？
\paragraph{Neural Network Training}
We further analyze the performance of B2Opt in the field of neuroevolution. We evaluate the performance of training a convolutional neural network \cite{AndrewHoward2017MobileNetsEC} using B2Opt on the MNIST classification task. The structure of this network model is shown in Table \ref{tab:mnist net}. This task involves a large number of parameters. B2Opt and baselines are tasked with solving the parameters of this neural network to maximize test accuracy. The network does not use any bias, and the total number of parameters is 567. If the SGD algorithm directly optimizes the neural network, the network can achieve about 90$\%$ test accuracy, which proves that the network architecture is effective. We use 1000 samples from the MINST dataset to form the training set and another 1000 samples as the test set.

\subsubsection{Baselines}
B2Opt is compared with the state-of-the-art BBO methods, including non-L2O and L2O-based.

\textbf{EA baselines}. DE(DE/rand/1/bin) \cite{das2010differential}, ES(($\mu$,$\lambda$)-ES), IPOP-CMA-ES \cite{1554902}, L-SHADE \cite{6900380}, and CMA-ES \cite{Hansen2016TheCE}, where DE \cite{das2010differential} and ES are implemented based on Geatpy \cite{geatpy}, CMA-ES and IPOP-CMA-ES are implemented by cmaes \footnote{https:
//github.com/CyberAgentAILab}, and L-SHADE is implemented by pyade \footnote{ https://github.com/xKuZz/pyade}. The reasons for choosing these EA baselines are the following:

\begin{itemize}
    \item DE(DE/rand/1/bin): A classic numerical optimization algorithm. 
    \item ES(($\mu$,$\lambda$)-ES): A classic variant of the evolution strategy.
    \item CMA-ES: CMA-ES is often considered the state-of-the-art method for continuous domain optimization under challenging settings (e.g., ill-conditioned, non-convex, non-continuous, multimodal).
    \item IPOP-CMA-ES: The state-of-the-art variant of CMA-ES.
    \item L-SHADE: The state-of-the-art variant of DE.
\end{itemize}

\textbf{Bayesian optimization}. Dragonfly \cite{Kirthevasan2020tuning} and SAASBO \cite{pmlr-v161-eriksson21a} are employed as a reference. Here are the reasons for choosing these two algorithms:
\begin{itemize}
    \item  Dragonfly: A representative algorithm for Bayesian optimization.
    \item  SAASBO: A state-of-the-art large-scale Bayesian optimization algorithm.
\end{itemize}

\textbf{L2O-based methods}. 
We chose three state-of-the-art L2O-based methods for comparison with B2Opt.
\begin{itemize}
    \item  L2O-swarm \cite{cao2019learning}: A representative L2O method for BBO. 
    \item  LES \cite{lange2022discovering}: A recently proposed learnable ES. It uses a data-driven approach to discover new ES with strong generalization performance and search efficiency.
    \item LGA \cite{lange2023discovering}: A recently proposed learnable GA that discovers new GA in a data-driven manner. The learned algorithm can be applied to unseen optimization problems, search dimensions, and evaluation budgets.
\end{itemize}
 
\textbf{B2Opt}. We design three B2Opt models, including \textit{3 OBs with WS}, \textit{5 OBs without WS}, and \textit{30 OBs with WS}. \textit{3 OBs with WS} represents that B2Opt has 3 OB modules, and these OBs share the weights of each other. Each OB consists of 1 SAC, 1 FM, and 1 RSSM.
In general, B2Opt represents 30 OBs with WS.

\subsubsection{Parameters}
Next, we show the parameter settings of these methods: 

\textbf{B2Opt}. In \textit{30 OBs with WS}, these 30 OBs share parameters. \textit{5 OBs without WS} has 5 OBs, and no parameters are shared among them. During the training process, B2Opt is trained for 1000 epochs. The initial learning rate ($lr$) was set to 0.01 and $lr$ = $lr \times 0.9$ every 100 cycles. The 2-norm of the gradient is clipped so that it is not larger than 10. The bias of the function is regenerated each epoch, and a new batch of random initial populations is generated.

\textbf{Baselines}. For all cases, we choose the optimal hyperparameters. For ES and DE, the population size is set to 100, $\lambda = 0.5$, $cr = 0.5$, $F = 0.5$, and the rest of the parameters are the default ones set by geatpy. For CMA-ES, the rest of the parameters take the default values of the package camaes. The population size of L-SHADE is 100, and the rest of the parameters adopt the default parameters of the package pyade. The initial population size of IPOP-CMA-ES is 30, and the population growth factor is 2. L-SHADE and IPOP-CMA-ES are known for their fast convergence. For all EA baselines, the maximum number of evaluations is set to 3100, which is consistent with the number of evaluations of B2Opt (30 OBs with WS). We set the maximum number of generations of Dragonfly and SAASBO to 100. Because the runtime overhead of these two Bayesian optimization algorithms is very expensive, it takes several days to complete 100 generations when $d = 100$. Even at $d=10$, it takes many hours to complete 100 evaluations. But B2Opt can finish running in 1 second, so the expensive actual running time overhead of SAASBO and Dragonfly is unacceptable. The population size of LGA and LES is 100, and the maximum number of function evaluations is 3100. We use the optimal trained model parameters officially provided by their project website\footnote{https://github.com/RobertTLange/evosax}. The $F^{Train}$ and $F^{test}$ is the same as that of B2Opt. In the test phase, L2O-Swarm was run to convergence.

To ensure validity, all experimental results were averaged to the optimal values of the five groups of experiments, each of which performed 64 runs. All experiments were performed on a Ubuntu 20.04 PC with Intel(R) Core I7 (TM) I3-8100 CPU at 3.60GHz and NVIDIA GeForce GTX 1060.

\begin{figure*}
\vskip 0.2in
\centerline{\includegraphics[width=0.9\linewidth]{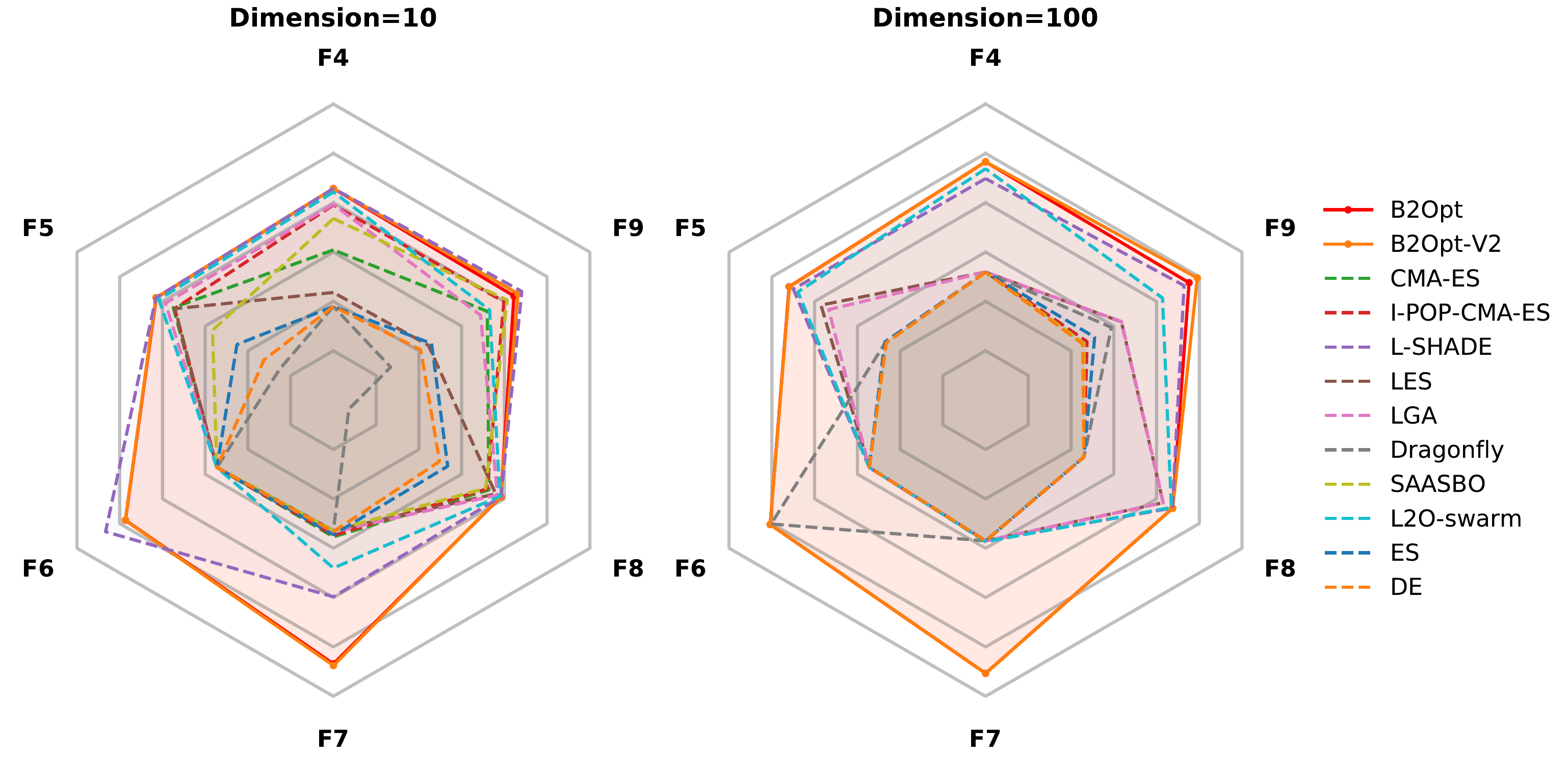}}
\caption{The compared results of all methods on six functions. The structure of B2Opt is \textit{30 OBs with WS}. The left schematic diagram and the right schematic diagram are the test results when the dimensions are 10 and 100, respectively. In order to present the experimental results more intuitively, we mark the experimental results as $a$, and the figure shows $\exp(-a)$ normalized by z-score. Larger values represent better performance.}
% \vskip -0.2in
\label{fig:radar}
\end{figure*}

\subsection{Results}
\begin{figure*}[!t]
 \centering
 \subfloat[F4]{\includegraphics[width=2.4in]{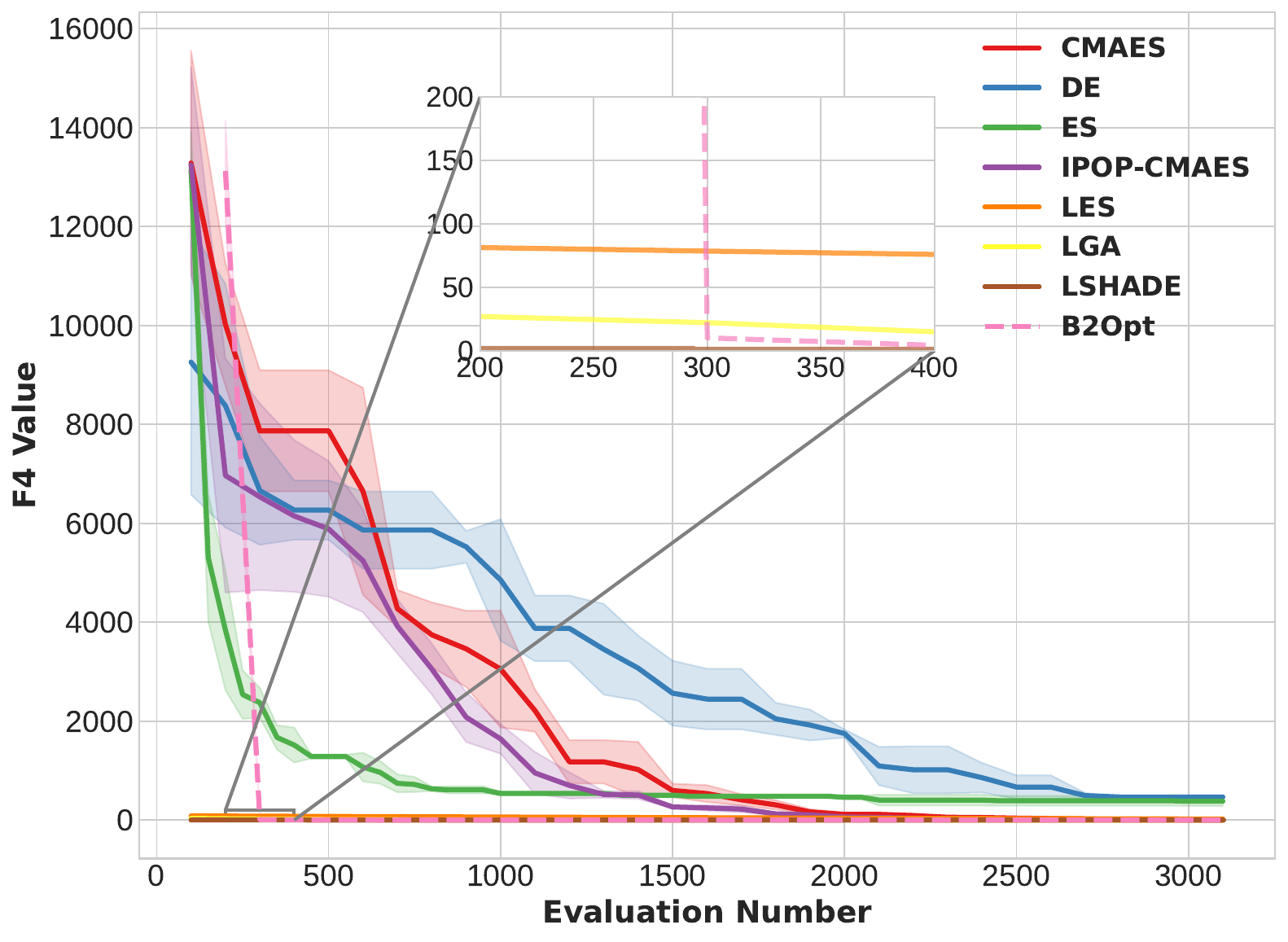}}
 \subfloat[F5]{\includegraphics[width=2.4in]{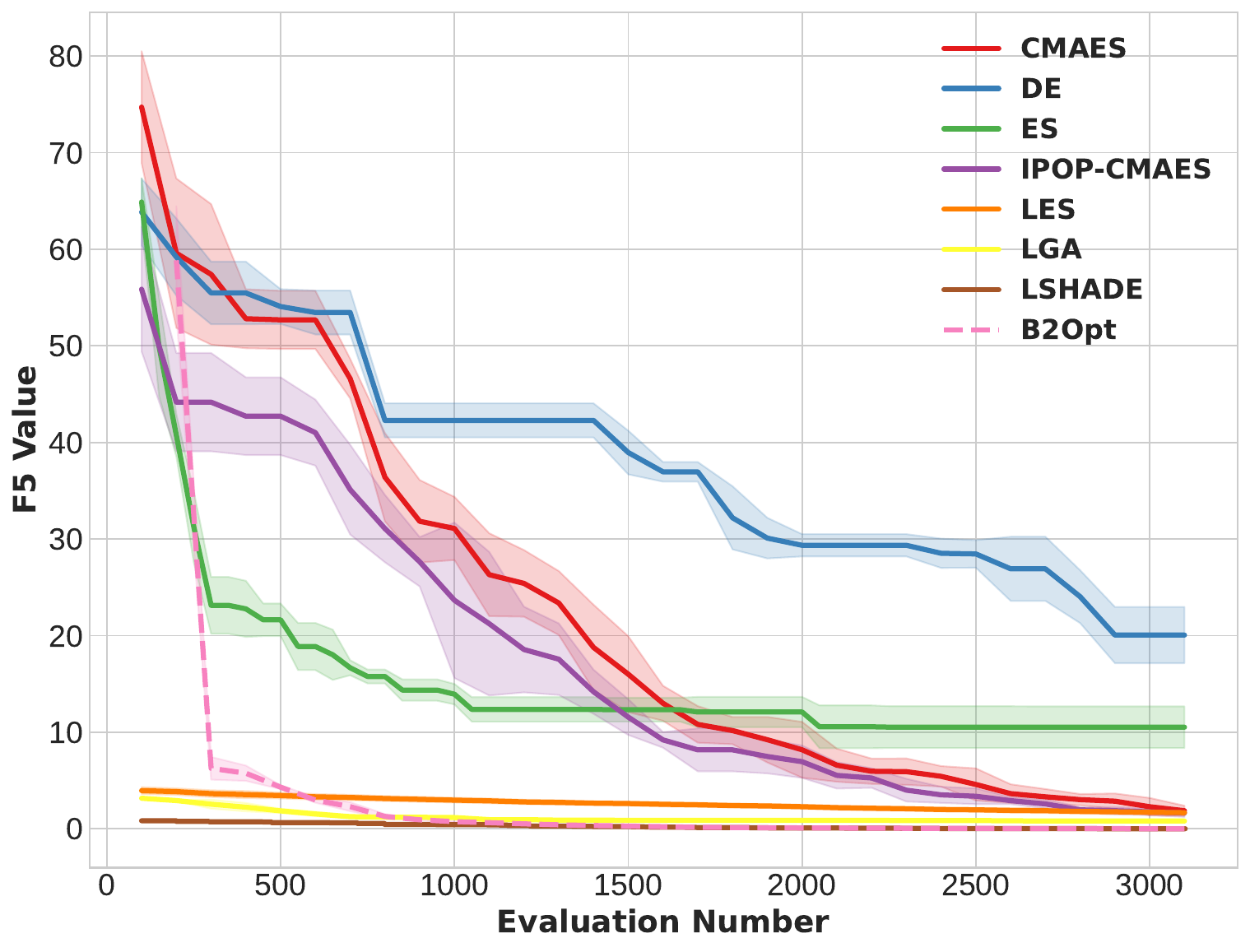}}
 \subfloat[F6]{\includegraphics[width=2.4in]{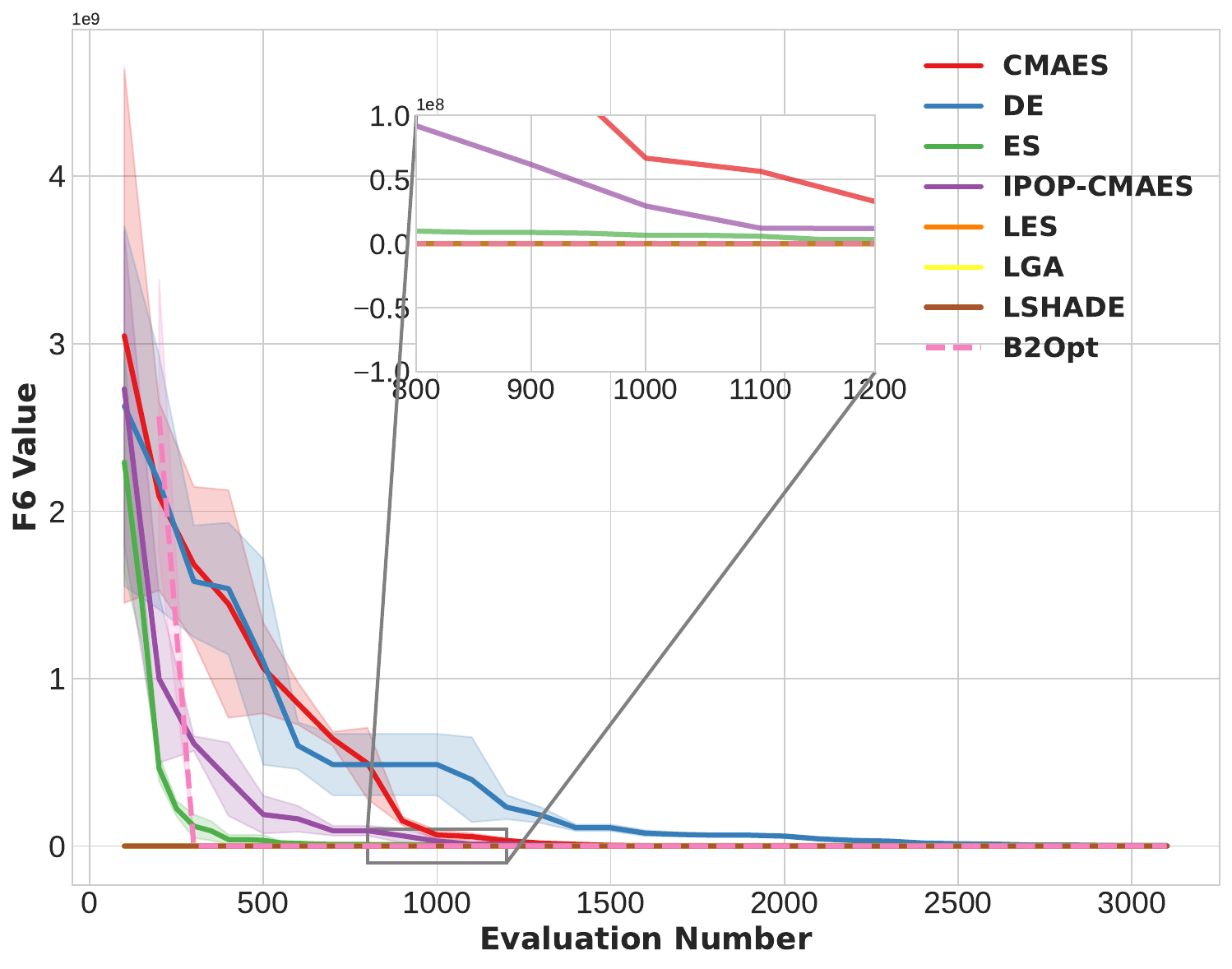}}\\
  \subfloat[F7]{\includegraphics[width=2.4in]{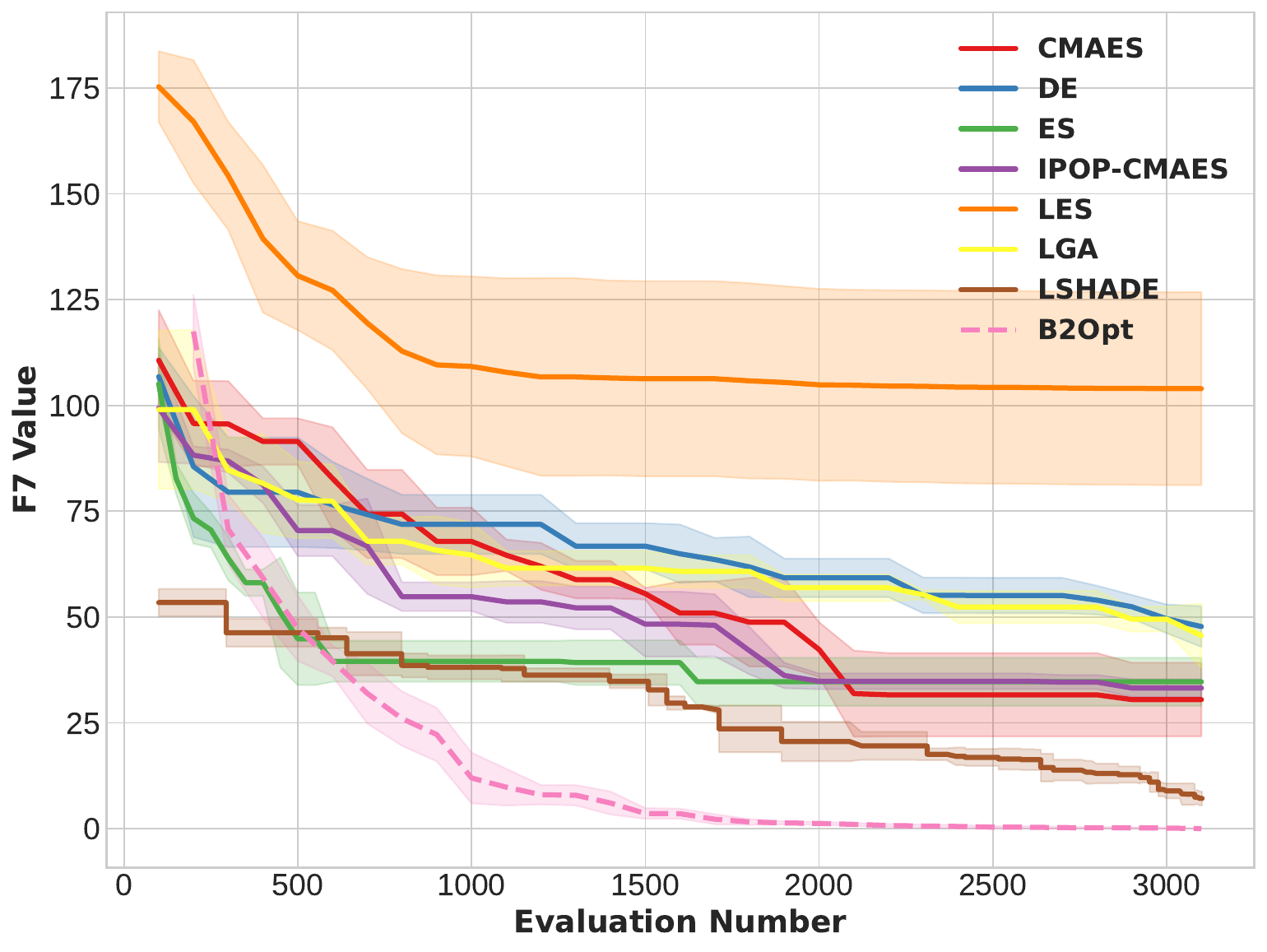}}
 \subfloat[F8]{\includegraphics[width=2.4in]{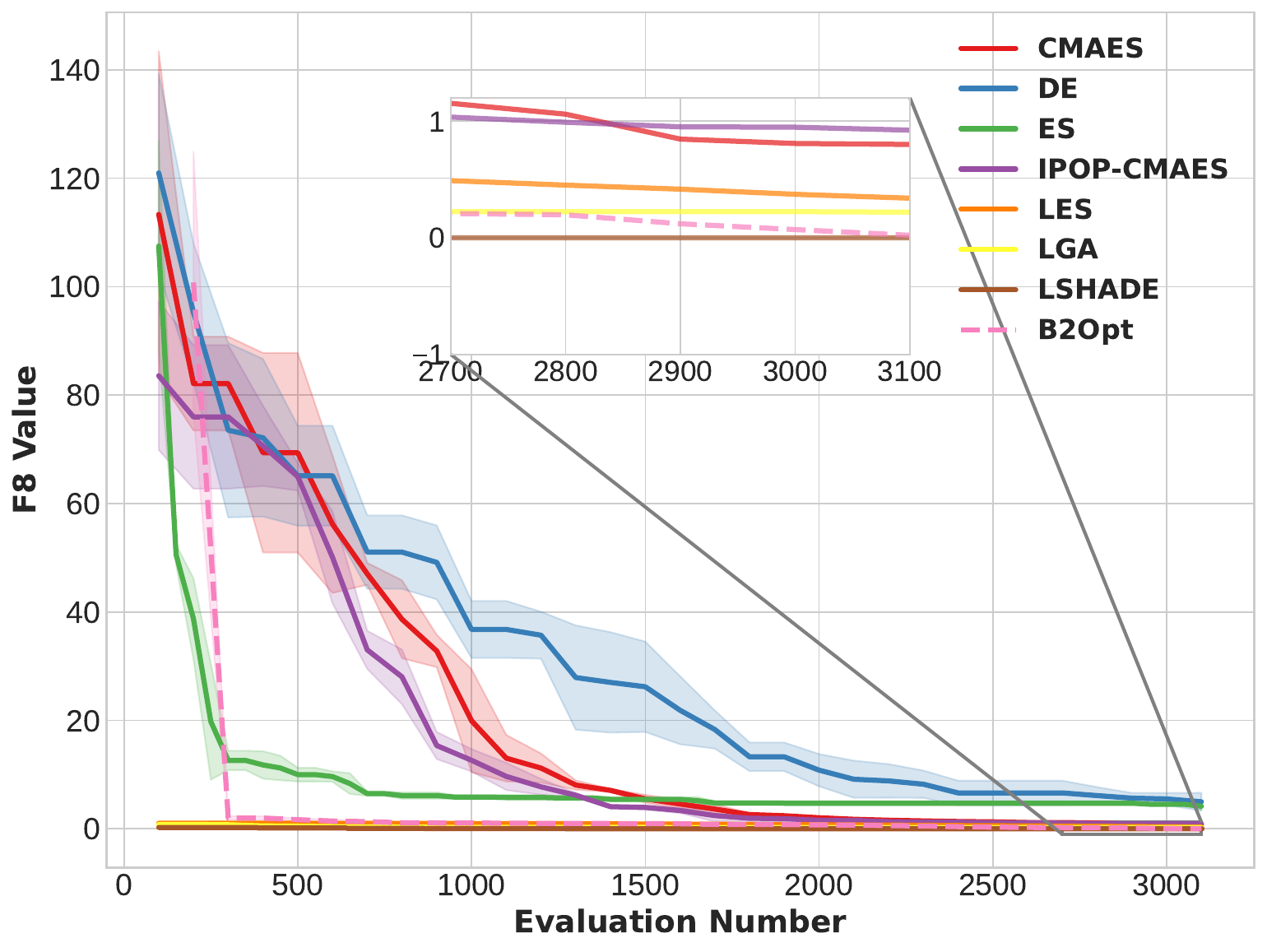}}
 \subfloat[F9]{\includegraphics[width=2.4in]{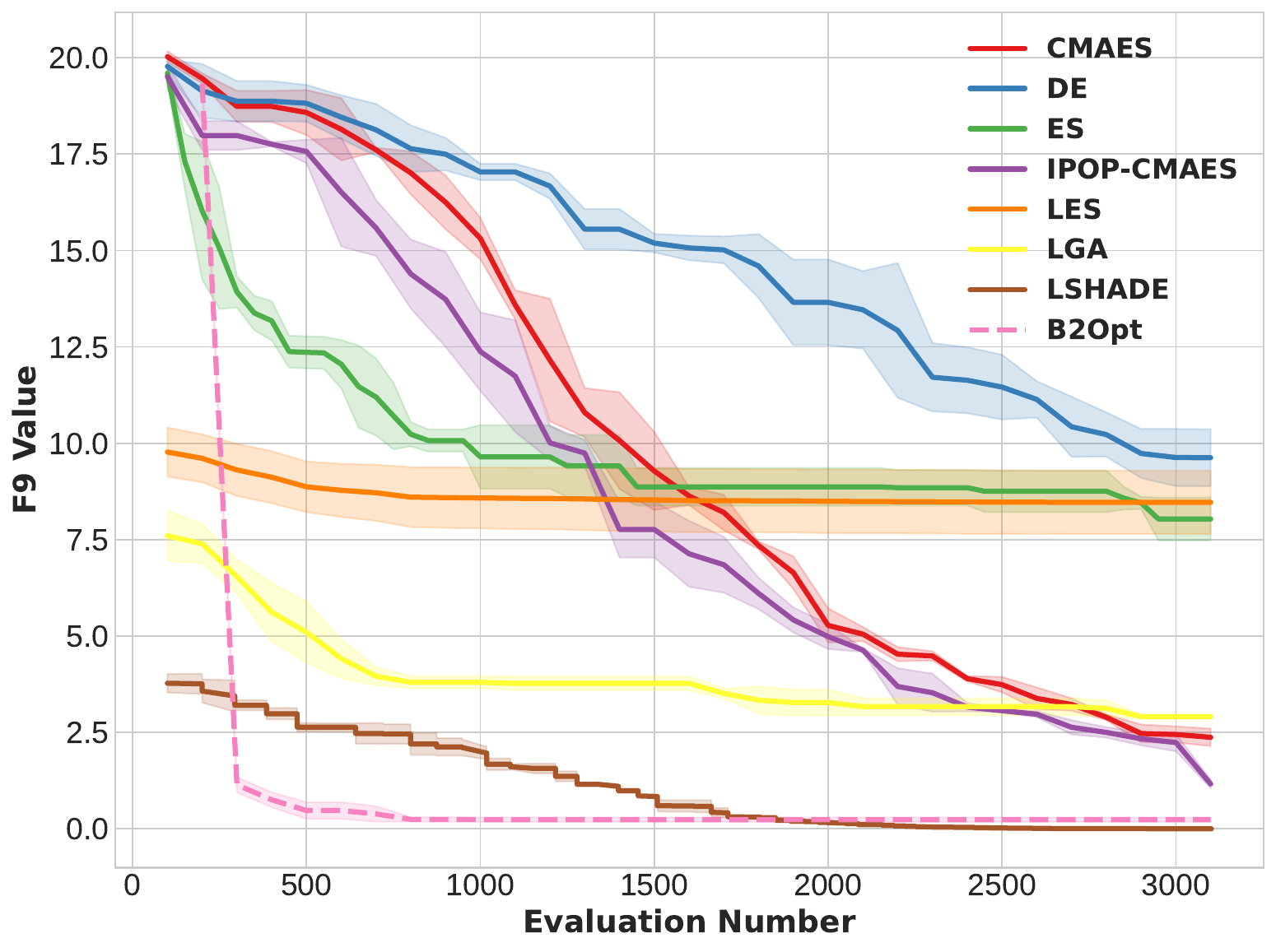}}
\caption{The convergence curves of B2Opt and other baselines. It shows the convergence curve of these algorithms on functions in Table \ref{table:test} when $d=10$.}
\label{fig:trail_d10}
\end{figure*}

\begin{figure*}[!t]
 \centering
 \subfloat[F4]{\includegraphics[width=2.4in]{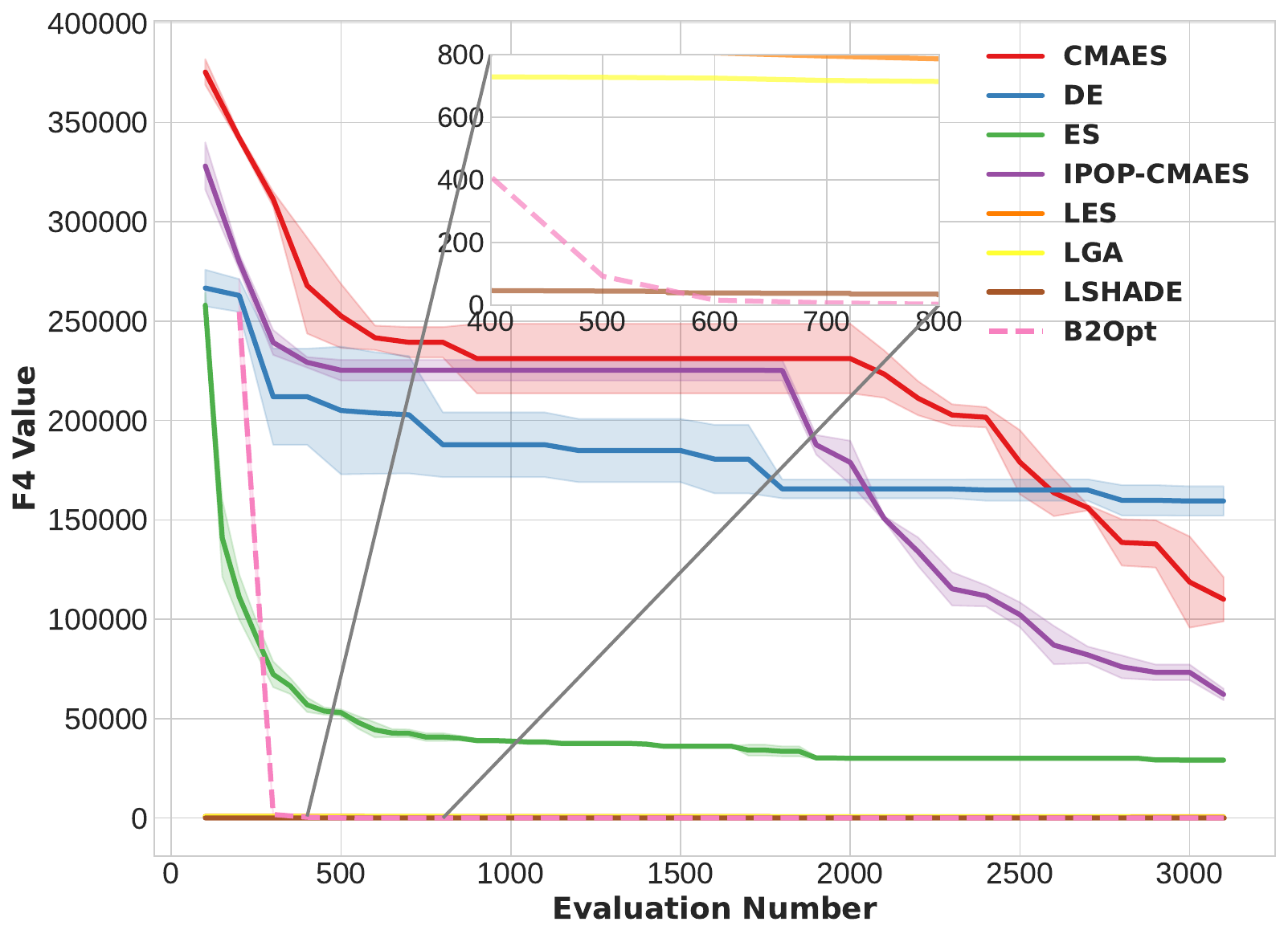}}
 \subfloat[F5]{\includegraphics[width=2.4in]{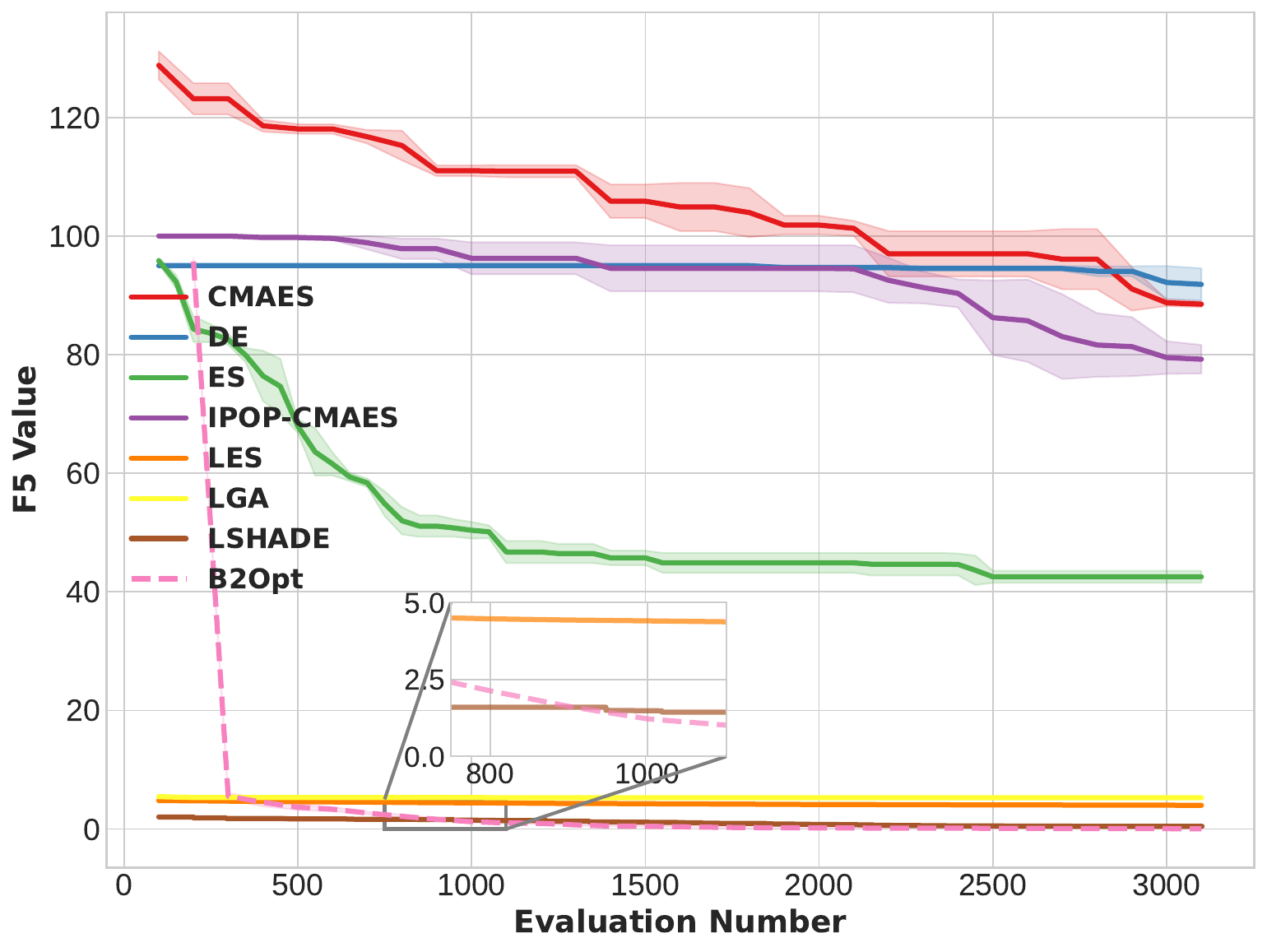}}
 \subfloat[F6]{\includegraphics[width=2.4in]{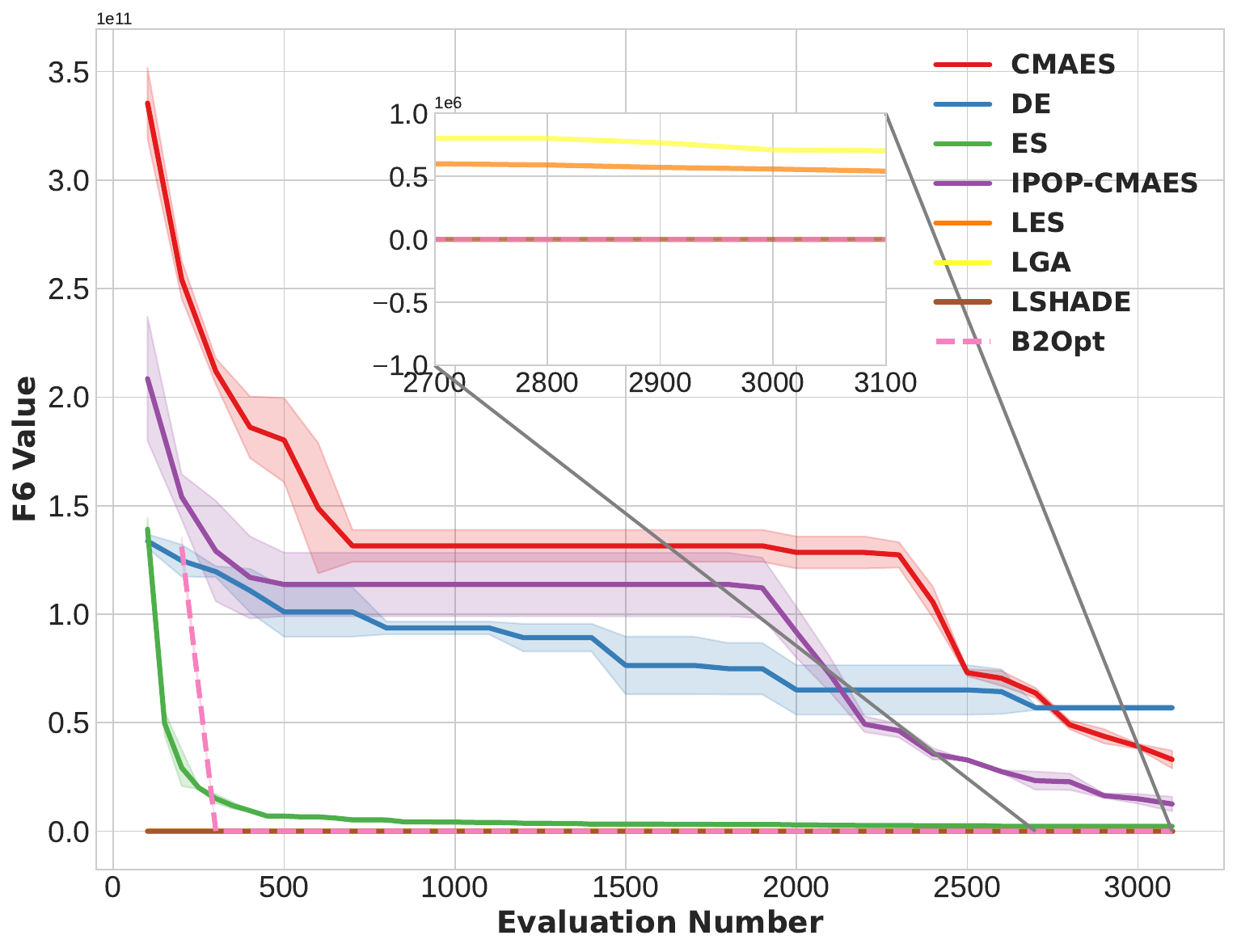}}\\
  \subfloat[F7]{\includegraphics[width=2.4in]{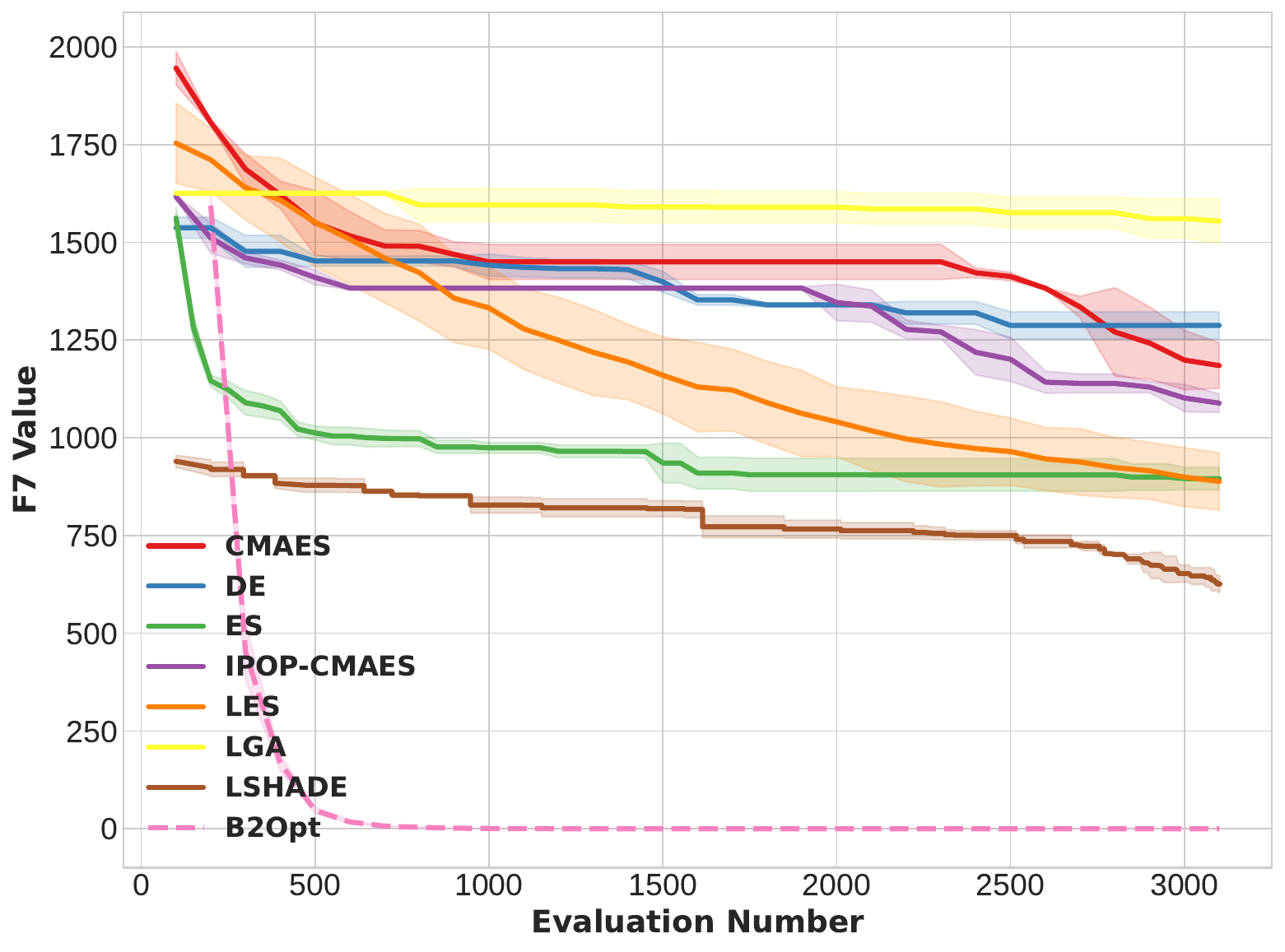}}
 \subfloat[F8]{\includegraphics[width=2.4in]{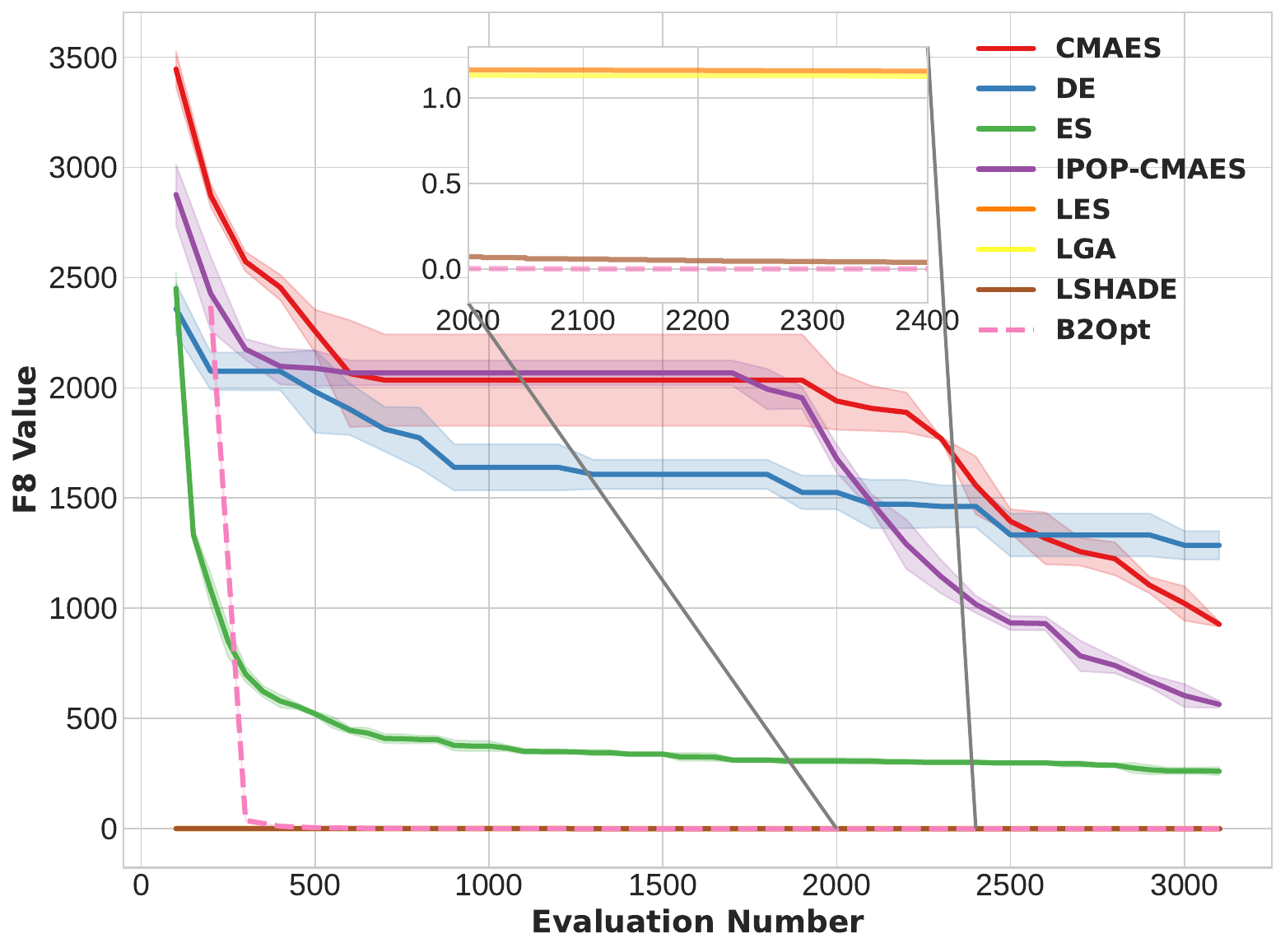}}
 \subfloat[F9]{\includegraphics[width=2.4in]{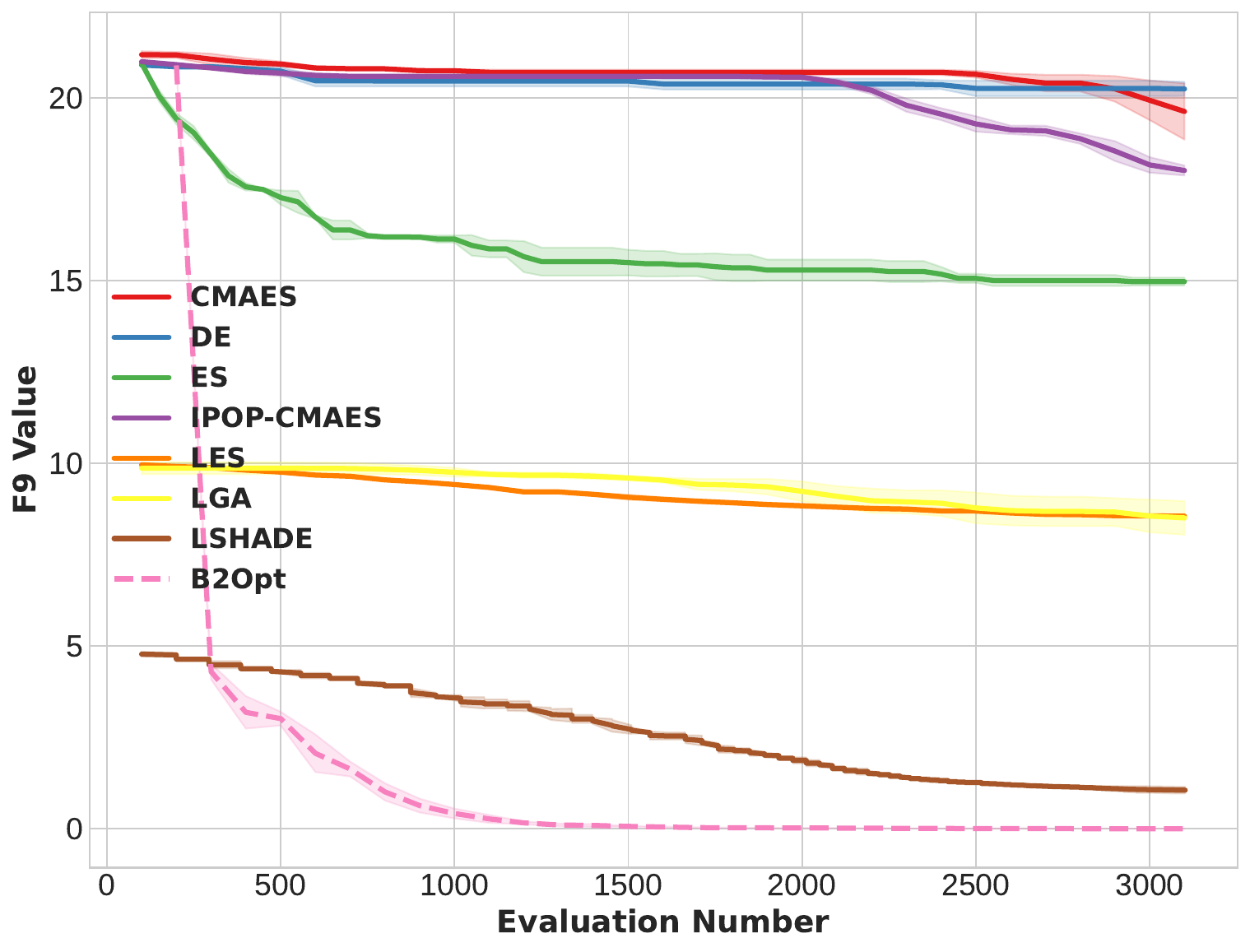}}
\caption{The convergence curves of B2Opt and other baselines. It shows the convergence curve of these algorithms on functions in Table \ref{table:test} when $d=100$.}
\label{fig:trail_d100}
\end{figure*}

\paragraph{Synthetic Functions}
Here, the structure of B2Opt is \textit{30 OBs with WS}. The results on synthetic functions are provided in Fig. \ref{fig:radar}. We also plot the convergence curves of all methods on F4-F9 with $d=\{10,100\}$, as shown in Figs. \ref{fig:trail_d10} and \ref{fig:trail_d100}. B2Opt converges quickly and can obtain better solutions with little budget. The fewer function evaluations, the greater the advantage of B2Opt. When $d=10$, B2Opt achieved better performance than all SOTA methods except for L-SHADE in all cases. B2Opt loses five cases to L-SHADE and outperforms L-SHADE in one case (F7). L-SHADE is the optimal DE variant, which is heavily engineered, and intricately hand-designed by researchers combining a wealth of expert knowledge. It has been iteratively updated many times over the years. L-SHADE can obtain high-performance initialized populations on F4-F9, far better than B2Opt's. However, the convergence speed of B2Opt is much faster than that of L-SHADE, and B2Opt is highly competitive, especially with a minimal number of function evaluations. However, when $d=100$, B2Opt outperforms all SOTA methods in F4-F9. When the problem dimension is 10, the problem is relatively simple, and when the problem dimension is 100, the problem becomes very complicated. B2Opt achieves the best performance on all tested functions in complex cases. L-SHADE has strong advantages over simple problems. Note that the hyperparameters of L-SHADE are optimally tuned to the performance of the target task. However, B2Opt is only trained on the low-fidelity surrogate functions F1-F3 of the target task.

LES and LGA are the latest learnable evolutionary algorithms with powerful performance. They are trained on a large number of BBOB functions, many of which are the same or similar to the test functions shown in Table \ref{table:test}. For example, the LES and LGA training datasets include F4, F5, F6, and F8. As shown in Figs. \ref{fig:trail_d10} and \ref{fig:trail_d100}, the initial population have achieved good performance. B2Opt only performs simple training on F1, F2, and F3 in Table \ref{table:test}, and F4-F9 is unseen for B2Opt in the training stage. However, B2Opt outperforms LES and LGA in all tested functions.

%51*10=510
B2Opt-V2 outperforms B2Opt overall. However, B2Opt-V2 performs worse than B2Opt on F8. B2Opt-V2 is trained on a high-fidelity surrogate function of the target task. We found that F8 has a "Rotated" feature and a wide range of $z_i$ values, which cause B2Opt-V2 to be overfitted. B2Opt-V2 outperforms B2Opt after we increase the training sample size.

These cases also show the excellent generalization ability of B2Opt on more tasks unseen during the training stage. The transferability of B2Opt is proportional to the fitness landscape similarity between the training set and the problem. Although new problem attributes are unavailable in the training set, B2Opt can perform better. However, this conclusion only holds when the similarity between the problem and training dataset is high. Now we further analyze the detailed reasons for the excellent effect of B2Opt.
%%%%discuss

We train B2Opt based on F1-F3 and make it successfully optimize F4-F9, and we also analyze the reason for this case.

\textbf{From the expression of the functions}, F1-F3 are low-fidelity surrogate functions of the target functions F4-F9. F1-F3 contain the information of objective functions (F4-F9). For example, F7 can be decomposed into $\sum_{i=1}^{D} z_i^2-\sum_{i=1}^{D}10\cos(2\pi z_i)+ \sum_{i=1}^{D}10$. F2 is the low-fidelity surrogate function of $\sum_{i=1}^{D} z_i^2$. F1 is the low-fidelity surrogate function of $\sum_{i=1}^{D}10\cos(2\pi z_i)$. For other functions in F4-F9, we can find similar surrogate functions from F1-F3. B2Opt can use this information to maximize the matching degree between the learned optimization strategy and the objective function. However, F6 is less similar to F1-F3 than F4, F5, and F7-F9. Herefore, although the performance of B2Opt on F6 is better than that of the comparison algorithm, it still needs improvement.

\begin{table*}[htbp]
%\vskip 0.15in
\caption{The results of all baselines on the planar mechanical arm. The structure of B2Opt is \textit{30 OBs with WS}. Simple Case (SC): searching for different angles with fixed lengths. Complex Case (CC): searching for different angles and lengths. w/t/l represents win/tie/loss to B2Opt.}
\begin{center}
%\resizebox{\textwidth}{!}{
    \begin{tabular}{cccccccccc|c|c}
    \hline
    & $r$     & DE    & ES    & CMA-ES & L2O-Swarm &L-SHADE&I-POP-CMA-ES&LGA&LES & B2Opt & \textit{Untrained}\\
    \hline
    \multirow{3}[2]{*}{SC} & 100   &1.60(0.08)&10.7(0.40) & 1.25(0.05)& 40.4(3.89)&0.66(0.03)&1.17(0.02)&540(1.97)&546(8.22)&\textbf{0.38(0.01)}&0.8(0.05)\\
          & 300   & 2.65(0.08) & 35(2.41) &1.84(0.07) & 69.5(3.77)&0.54(0.02)&1.21(0.06)&562(2.1)&572(4.92)&\textbf{0.52(0.02)}& 4.26(0.33) \\
          & 1000  & 155(1.45) & 153(2.19) & 154(2.31)  & 176(7.20)& \textbf{0.4(0.03)}&36.8(1.00)&716(1.74)&719(4.03)
 & 16.9(0.30)& 126(0.85)\\
    \hline
    \multirow{3}[2]{*}{CC} & 100   & 1.17(0.03) &9.22(0.52)& 0.78(0.02) & 31.9(1.78)&0.25(0.02)&0.72(0.02)&108(1.34)&223(6.51) & \textbf{0.17(0.01)} & 0.94(0.04)\\
          & 300   & 13.6(0.53) & 38.3(0.60)&4.77(0.37)& 89.1(1.96)&54.5(5.23)&0.92(0.02)&161(2.67)&259(4.29)&\textbf{0.22(0.01)}& 18.6(0.39)\\
          & 1000  & 278(1.96) &235(1.41) &239(0.83)& 262(2.99)&218(0.89)&	167(1.15)&421(0.93) &493(1.15)& \textbf{14.7(0.70)}& 214(0.93)\\
    \hline
     \multicolumn{2}{c}{w/t/l} &0/0/6&0/0/6& 0/0/6& 0/0/6&1/0/5 &0/0/6&0/0/6& 0/0/6 &- & 0/0/6 \\
    \hline
    \end{tabular}
    %}
  \label{tab:pma1}
\end{center}
%\vskip -0.1in
\end{table*}

\textbf{From the perspective of landscape features}, F1-F3 include the following features: unimodal, multimodal, separable, and non-separable. The landscape features included in F4-F9 are as follows:
\begin{itemize}
    \item F4: Unimodal, Separable
    \item F5: Unimodal, Separable
    \item F6: Multimodal, Non-separable, Having a very narrow valley from local optimum to global optimum, Ill-conditioned
    \item F7: Multimodal, Separable, Asymmetrical, Local optima’s number is huge
    \item F8: Multi-modal, Non-separable, Rotated
    \item F9: Multi-modal, Non-separable, Asymmetrical
\end{itemize}

The landscape features of F4 and F5 can be found in F1-F3. F6-F9 all have new features. The interference strength of different characteristics to landscape is arranged as follows: \textbf{Having a very narrow valley from local optimum to global optimum}$>$\textbf{Asymmetrical, Local optima’s number is huge}$>$\textbf{Asymmetrical}$>$\textbf{Rotated}. Therefore, B2Opt has the best generalization performance on F4, F5, F8, the second-best generalization performance on F7 and F9, and the worst on F6.

\begin{table}[htbp]
  \centering
  \caption{The designed convolution module to replace SAC. one represents one reflection padding.}
    \begin{tabular}{ccccc}
    \toprule
    Layer ID & layer type & padding & stride & kernel size \\
    \midrule
    1     & depth-wise convolution & one & 1     & 567$\times$3$\times$3 \bigstrut\\
    2     & ReLu  & \XSolidBrush  &  \XSolidBrush &  \XSolidBrush \bigstrut\\
    3     & depth-wise convolution & one & 1     & 567$\times$3$\times$3 \bigstrut\\
    4     & ReLu  &  \XSolidBrush  &  \XSolidBrush &  \XSolidBrush \bigstrut\\
    5     & depth-wise convolution & one & 1     & 567$\times$3$\times$3 \bigstrut\\
    6     & ReLu  &  \XSolidBrush  &  \XSolidBrush &  \XSolidBrush \bigstrut\\
    7     & depth-wise convolution & one & 1     & 567$\times$3$\times$3 \bigstrut\\
    \bottomrule
    \end{tabular}%
  \label{tab:cnn in place of sac}%
\end{table}

\paragraph{Planner Mechanic Arm}
The detailed experimental results are in Tables \ref{tab:pma1}. Note that the parentheses in the table show the standard deviation if not otherwise specified. The structure of B2Opt is \textit{30 OBs with WS}. \textit{Untrained} represents the untrained B2Opt.  We randomly selected 600 target points within the range of $r \leq 1000$ to form a set $S$, where $r$ represents the distance from the target point to the origin of the mechanic arm, as shown in Fig. \ref{fig:pma}. During the training process of B2Opt, a sample point set $s$ is re-extracted from $S$ for training every $T$ training cycle. In the testing process, we extracted 128 target points ($S^{test}$) in the range of $r \leq 100$, $r \leq 300$, and $r \leq 1000$, respectively, for testing. The purpose of testing in three different regions is to explore the generalization performance of B2Opt further. %We evaluate the generalization ability of the algorithm by $\left(\sum_s^{S^{test}} f(L, \alpha, s)\right) / |S^{test}|$.
B2Opt wins all SOTA methods and achieves the best results in comparison with other algorithms. B2Opt loses once to L-SHADE in SC with $r=1000$. However, in the other 5 cases, B2Opt outperforms L-SHADE. In particular, on the complex problem (CC), B2Opt is more dominant and stable than L-SHADE. The optimization objective of this problem does not exist in the training sets of LGA and LES. A sharp performance degradation of LGA and LES can be observed. We also find the surprising phenomenon that \textit{Untrained} outperforms most of the baselines, which suggests that the randomly initialized B2Opt also possesses some ability to produce and select potential solutions.

\begin{table*}[htbp]
  \centering
  \caption{The classification accuracy of all methods on the MNIST dataset. Datasize represents the proportion of data sets involved in training.}
  \label{table:mnist}%
\begin{tabular}{c|cccccccc}  
    \toprule
    Datasize & B2Opt & ES    & DE    & CMA-ES & I-POP-CMA-ES& L-SHADE&LGA &LES \\
    \midrule
    0.25  & \textbf{0.69(0.01)} & 0.21(0.02) & 0.21(0.01) &0.57(0.04)&0.33(0.02)&0.31(0.03)&0.53(0.02)&0.14(0.02)
\\
    0.5   &\textbf{0.80(0.01)} & 0.21(0.01) & 0.24(0.01) & 0.61(0.03)&0.37(0.04)&0.29(0.01)&0.51(0.01)&0.25(0.03)
\\
    0.75  & \textbf{0.85(0.00)} & 0.21(0.01) & 0.22(0.02) &0.61(0.03)&0.45(0.02)&0.30(0.07)&0.52(0.05)&0.23(0.07)
\\
    1     & \textbf{0.86(0.00)} & 0.25(0.02) & 0.23(0.02) & 0.67(0.01)&0.42(0.02)&0.30(0.01)&0.54(0.04)&0.28(0.01)
 \\
    \bottomrule 
\end{tabular}
\end{table*}

\begin{figure}[htbp]
  \begin{center}
\centerline{\includegraphics[width=0.9\columnwidth]{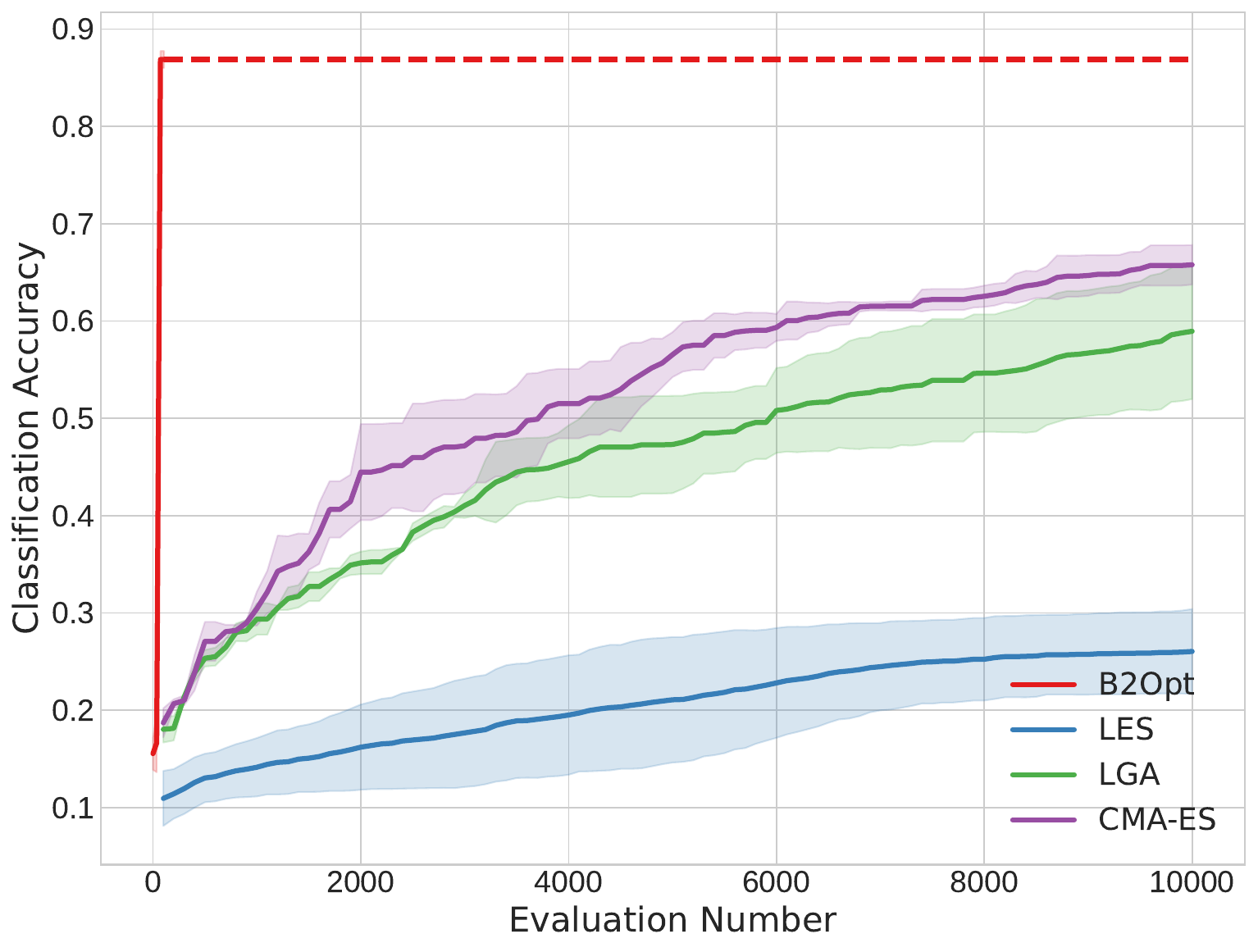}}
  \caption{The convergence curves of B2Opt and baselines on the task of neural network training. The X-axis represents the number of evaluations, and the Y-axis represents the test accuracy. B2Opt stops iterating when the number of evaluations is 144, and the red dotted line is for convenient comparison.}
  \label{fig: convergence mninst}
  \end{center}
\end{figure}

\paragraph{Neural Network Training}
The detailed results are shown in Table \ref{table:mnist}. The evaluation metric is test accuracy on the test set. While training B2Opt, the optimization objective of B2Opt is to minimize the cross-entropy loss, which is a surrogate function for metric accuracy. However, in the testing stage, the optimization goal of B2Opt and other baselines is to maximize the accuracy of the training set. We select 25\%, 50\%, 75\%, and 100\% data from the training set for training, respectively, constituting surrogate problems with different fidelity levels. B2Opt has 3 OBs that do not share weights. We replaced SAC with a lightweight convolution module with the structure shown in Table \ref{tab:cnn in place of sac}. This substitution is made because we expect B2Opt to perform more stably during training for this task. The population size of B2Opt is 36, meaning its number of evaluations is $36\times4=144$. The maximum number of evaluations for L-SHADE and I-POP-CMA-ES is 3000, which is $3000/144\approx 21$ times that of B2Opt. LGA and LES have a maximum number of generations of 100, and they are evaluated $10000/144\approx 69$ times as many times as B2Opt. Even in this unfair case, B2Opt achieves the best results for all fidelity levels. Fig. \ref{fig: convergence mninst} shows the convergence curve of B2Opt, LES, LGA, and CMA-ES on this task. B2Opt can achieve the best solution with the least number of evaluations.

\subsection{Parameter Analysis}
We analyze the effect of the deep structure, learning rate, and weight sharing between OBs on B2Opt. 

\subsubsection{B2Opt Architectures} We consider the performance of different B2Opt architectures. The experimental results are shown in Table \ref{table:paremeter}. They were sorted from good to worst based on their performance, and the result is \textit{30 OBs with WS}$>$\textit{5 OBs without WS}$>$\textit{3 OBs with WS}. Deep architectures have better representation capabilities and also lead to better performance. However, training non-weight-sharing B2Opts with more layers is challenging due to the difficulty of training deep architectures. \textit{Untrained} represents that the parameters of \textit{5 OBs without WS} are randomly initialized. \textit{5 OBs without WS} outperforms \textit{Untrained}, which demonstrates the effectiveness of the designed training process. 
\begin{table}[ht]
%\vskip 0.15in
\caption{The performance of different B2Opt structures.}
\label{table:paremeter}
\begin{center}
    \begin{tabular}{c|cccc}
    \toprule
     $f$  & \textit{Untrained} & \textit{5 OBs} & \textit{30 OBs} & \textit{3 OBs} \\
       &  & \textit{without WS} & \textit{with WS} & \textit{with WS} \\
    \midrule
     F4  & 0.28(0.09) & 0.08(0.03) & \textbf{1.2e-4(5e-5)} & 8.16(3.44) \\
     F5  & 0.37(0.05)  & 0.15(0.03) & \textbf{0.008(0.002)} & 1.47(0.40) \\
     F6  & 45.8(16.9)  & 15.43(2.34) & \textbf{8.93(0.03)} & 1891(1396) \\
     F7 & 1.08(0.72)   & 4.43(1.82) & \textbf{0.01(0.03)} & 35.72(8.52) \\
     F8   & 0.69(0.09) & 0.06(0.03) & \textbf{1e-5(2e-5)} & 0.82(0.10) \\
     F9  & 0.85(0.20)  & 0.29(0.07) & \textbf{0.01(0.003)} & 3.28(1.00) \\
    \bottomrule
    \end{tabular}
\end{center}
%\vskip -0.1in
\end{table}
We also find an interesting phenomenon: \textit{5 OBs without WS} outperforms \textit{3 OBs with WS} in all cases. Our untrained deep architecture, \textit{30 OBs with WS}, can achieve good results on simple planner mechanic arm problems, which shows that B2Opt retains the advantages of Transformer architecture and has strong generalization ability. We use the untrained \textit{5 OBs without WS} to test the complex planner mechanic arm problem, which performs poorly. 

We have observed that B2Opt can achieve better results with deeper architectures. However, it is currently difficult for us to train deep B2Opt. Moreover, as far as we know, the use of ES to optimize deep models has been studied a lot \cite{vicol2021unbiased}, which will be an essential research prospect in the future.

\begin{table*}[ht]
  \centering
  \caption{The effect of learning rate on B2Opt.}
    \begin{tabular}{ccccccc}
    \toprule
    $lr$    & F4    & F5    & F6    & F7    & F8    & F9 \bigstrut\\
    \hline
    \multicolumn{7}{c}{\textit{5 OBs without WS}} \bigstrut\\
    \hline
    0.1   & 0.93(7.42) & 0.31(0.49) & 2.04e7(2.03e8) & 15.3(6.4) & 0.36(0.16) & 0.61(0.18) \bigstrut\\
    0.01  & \textbf{0.01(0.003)} & \textbf{0.05(0.02)} & \textbf{9.57(0.22)} & \textbf{1.62(0.60)} & \textbf{0.03(0.01)} & \textbf{0.06(0.03)} \\
    0.001 & 0.88(3.41) & 0.36(0.12) & 226(1750) & 6.18(2.66) & 0.56(0.16) & 1.36(0.36) \\
    0.0001 & 0.06(0.03) & 0.13(0.03) & 13.6(2.11) & 0.83(0.50) & 0.17(0.04) & 0.28(0.10) \bigstrut\\
    \hline
    \multicolumn{7}{c}{\textit{30 OBs with WS}} \bigstrut\\
    \hline
    0.1   & 1.64(1.19) & 0.85(1.59) & 493(3110) & 28.4(4.35) & 0.47(0.11) & 2.82(0.5) \bigstrut\\
    0.01  & 0.05(0.30) & 0.09(0.03) & 39.2(240) & 1.05(1.40) & 0.01(0.06) & 0.28(0.07) \\
    0.001 & \textbf{1.01e-3(0.001)} & 0.02(0.01) & 9.03(0.18) & 0.03(0.02) & \textbf{0.003(0.001)} & 0.03(0.01) \\
    0.0001 & 1.50e-3(0.001) & \textbf{0.016(0.004)} & \textbf{9.01(0.13)} & \textbf{0.02(0.02)} & 0.006(0.002) & \textbf{0.02(0.0.01)} \bigstrut\\
    \hline
    \multicolumn{7}{c}{\textit{3 OBs with WS}} \bigstrut\\
    \hline
    0.1   & 3.04(5.55) & 0.98(0.4) & 1150(7930) & 43.3(8.87) & \textbf{0.64(0.12)} & 2.43(0.98) \bigstrut\\
    0.01  & 29.9(47.7) & 2.68(1.30) & 6.24e4(4.87e5) & 40.4(8.52) & 1.05(0.06) & 4.45(0.85) \\
    0.001 & 1.82(1.20) & 0.76(0.32) & 654(4780) & 7.00(7.11) & 0.79(0.13) & 1.91(0.53) \\
    0.0001 & \textbf{0.39(0.21)} & \textbf{0.33(0.07)} & \textbf{46.8(79.4)} & \textbf{2.22(2.41)} & 0.66(0.09) & \textbf{0.59(0.19)} \bigstrut\\
    \bottomrule
    \end{tabular}
  \label{tab:lr}%
\end{table*}%

\subsubsection{Learning Rate} We train B2Opt on the F1-F3 function set with different learning rates and then test it on the F4-F9 function set. The experimental results are shown in Table \ref{tab:lr}. \textit{5 OBs without WS} and \textit{30 OBs with WS} perform poorly when the learning rate is 0.1, which may be because the learning rate is too large, which affects the convergence of B2Opt during the training process. For \textit{5 OBs without WS}, setting the learning rate to 0.01 achieves the relatively best performance. A learning rate 0.0001 would be a good choice for \textit{30 OBs with WS} and \textit{3 OBs with WS}. However, our experiments are coarse-grained. The learning rate has a significant impact on B2Opt. Then using AutoML to search for the optimal hyperparameter combination of the model is expected to achieve better performance.

\subsubsection{Training Data Size} 
\begin{figure}[htbp]
  %\vskip 0.2in
  \begin{center}
\centerline{\includegraphics[width=0.85\linewidth]{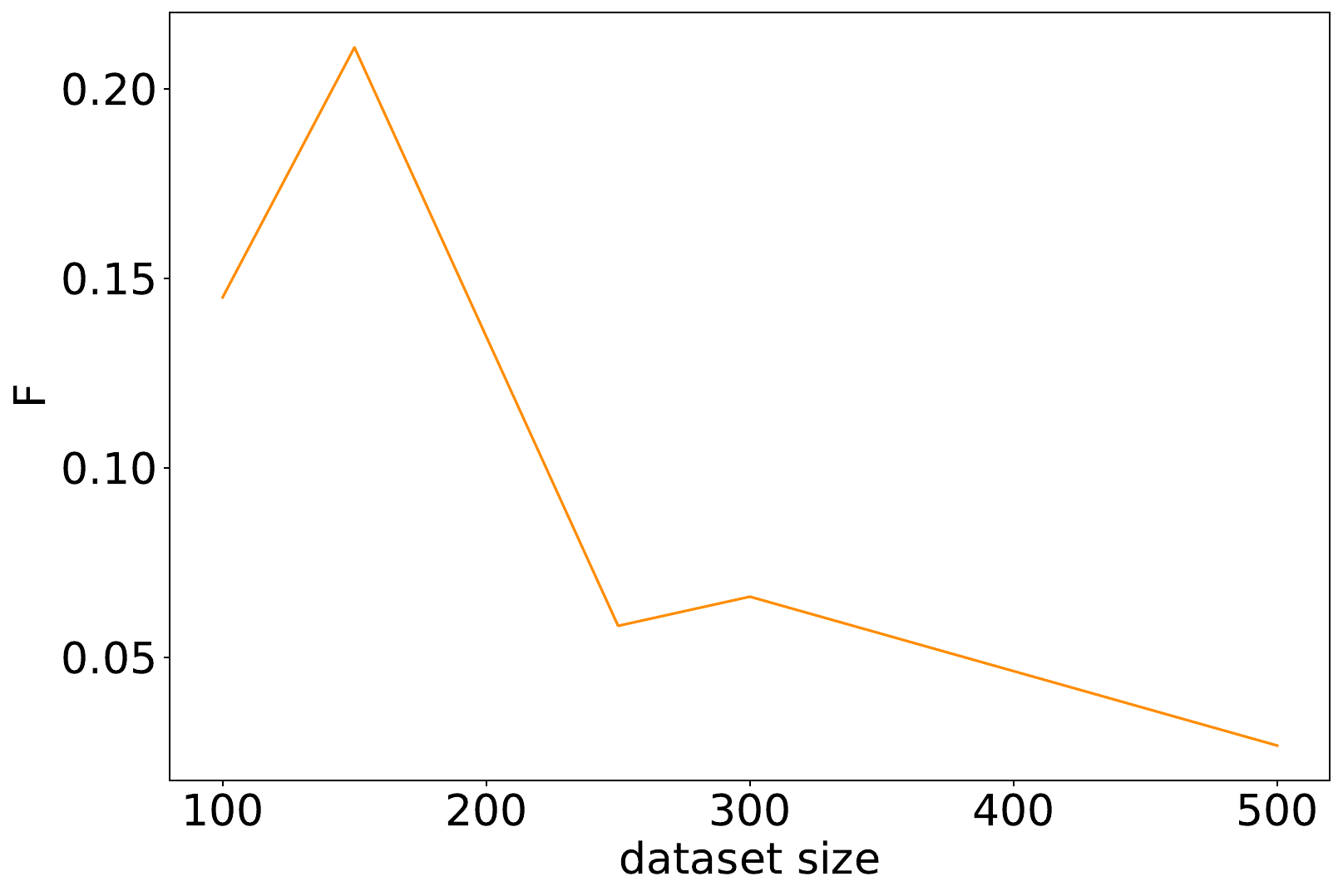}}
  \caption{The impact of training dataset size on B2Opt.}
  \label{fig:datasize}
  \end{center}
  %\vskip -0.2in
\end{figure}

We also explore the impact of the size of the training data set on the performance of the algorithm and take F4 as an example. It is trained on various biases of F4 and tested on F4 without bias. The training data set is $F^{train} = \{F4(x|\omega_{1,i}^{train}),\cdots, F4(x|\omega_{m,i}^{train})\}$.
Experimental results are shown in Fig. \ref{fig:datasize}, which show that the size of the training dataset significantly impacts the performance of B2Opt. As the amount of training data increases, the performance of B2Opt increases.

\begin{figure*}[htbp]
  %\vskip 0.2in
  \begin{center}
\centerline{\includegraphics[width=0.95\linewidth]{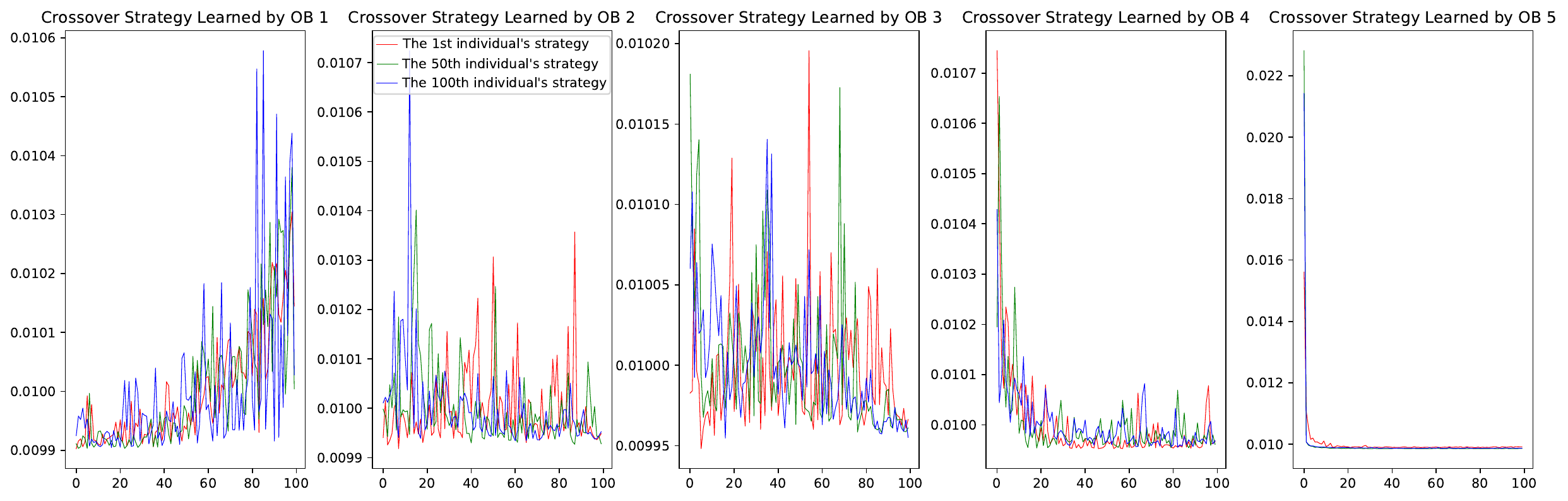}}
  \caption{Crossover Strategy learned by B2Opt.}
  \label{fig:crossover visualization}
  \end{center}
  %\vskip -0.2in
\end{figure*}

\subsection{Ablation Study}
This section considers the impact of different parts on B2Opt. We take B2Opt with 3 OBs and weight sharing as an example, which is trained on F1-F3 and tested on F4-F9. We remove SAC, FM, RSSM, and RC in B2Opt, respectively, and denote them as \textit{Not SAC}, \textit{Not FM}, \textit{Not RSSM}, and \textit{Not RC}. The experimental results are shown in Table \ref{table:ablation}. 
When their results are sorted from good to worst, the rank is B2Opt $>$ \textit{Not FM} $>$ \textit{Not RC} $\approx$ \textit{Not SAC} $\approx$ \textit{Not RSSM}. The role of FM is slightly weaker than that of the other three modules. Taken as a whole, the parts of SAC, RSSM, and RC are of equal importance. The absence of these core components can seriously affect the performance of B2Opt. At the same time, it also shows the effectiveness of the proposed four modules. The removal of any one of the modules in the crossover, mutation, and selection of EAs will degrade the performance of EAs. This shows that B2Opt implements a learnable EA framework that does not require human-designed parameters.

\begin{table}
\caption{The results of ablation study. $d$ = 10.}
\label{table:ablation}
\begin{center}
%\resizebox{\linewidth}{!}{
    \begin{tabular}{c|ccccc}
    \toprule
    $f$ & \textit{Not SAC} & \textit{Not FM} & \textit{Not RC} & \textit{Not RSSM} & B2Opt\\
    \midrule
   F4 & 52.7(15.0) & 23.3(6.68) & 50.5(6.81) & 271(346) & 3.03(5.82) \\
   F5 & 5.08(0.98) & 2.94(0.41) & 3.75(0.09) & 3.51(1.49) & 1.02(0.20) \\
   F6 & 1e5(9e4) & 1e4(1e4) & 5e4(1e4) & 7e6(1e8) & 4e3(7e4) \\
   F7 & 47.2(5.11) & 19.7(4.27) & 63.4(8.60) & 40.6(8.10) & 44.8(8.34) \\
   F8 & 1.18(0.04) & 1.19(0.07) & 0.87(0.07) & 3.28(1.14) & 0.60(0.17)\\
   F9 & 4.49(1.09) & 3.42(0.78) & 8.36(0.43) & 3.56(0.72) & 3.03(0.70) \\
    \bottomrule
    \end{tabular}
    %}
\end{center}
\end{table}

\begin{figure*}[!t]
 \centering
 \subfloat[]{\includegraphics[width=2.1in]{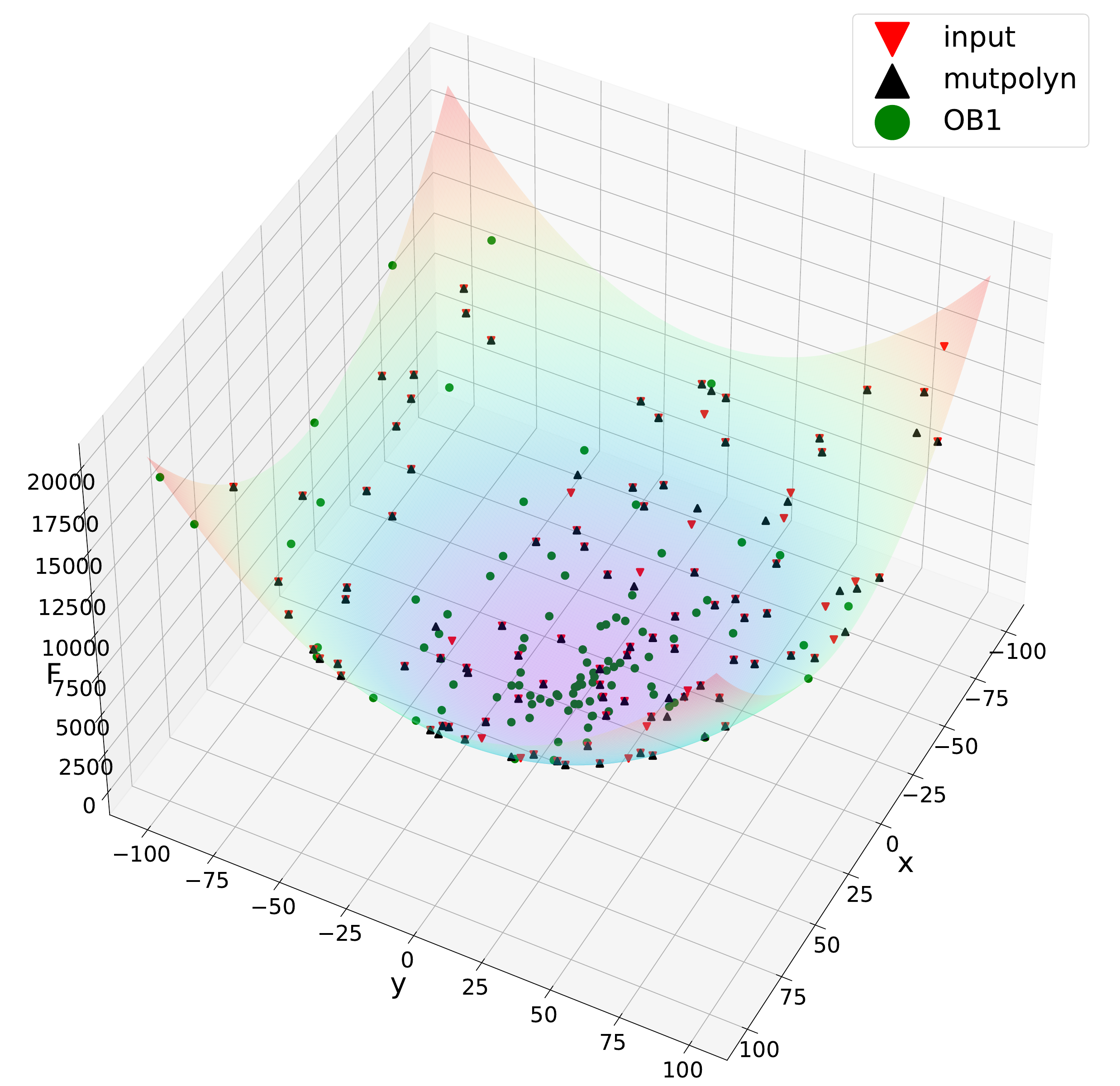}}
 \subfloat[]{\includegraphics[width=2.1in]{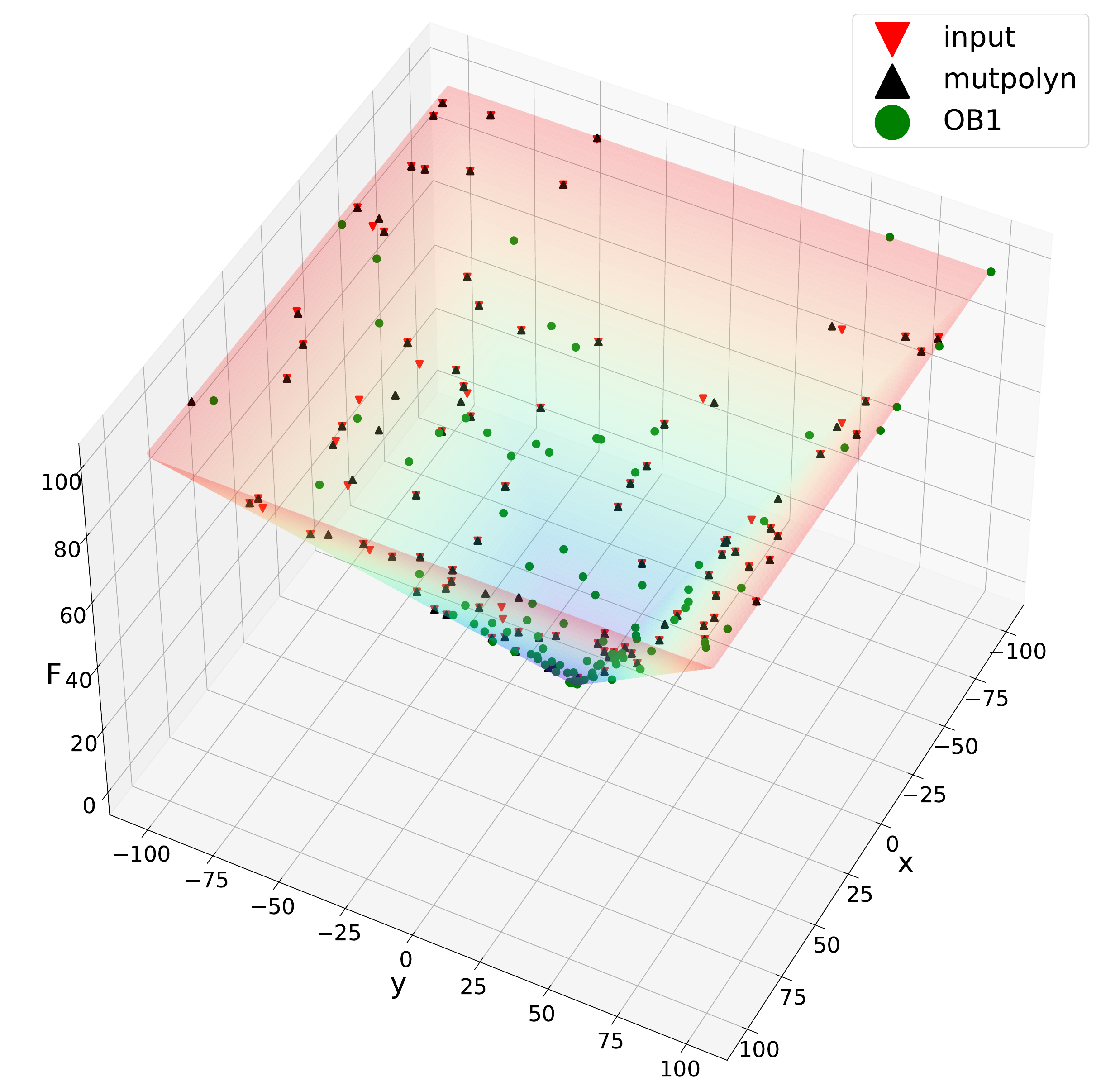}}
 \subfloat[]{\includegraphics[width=2.1in]{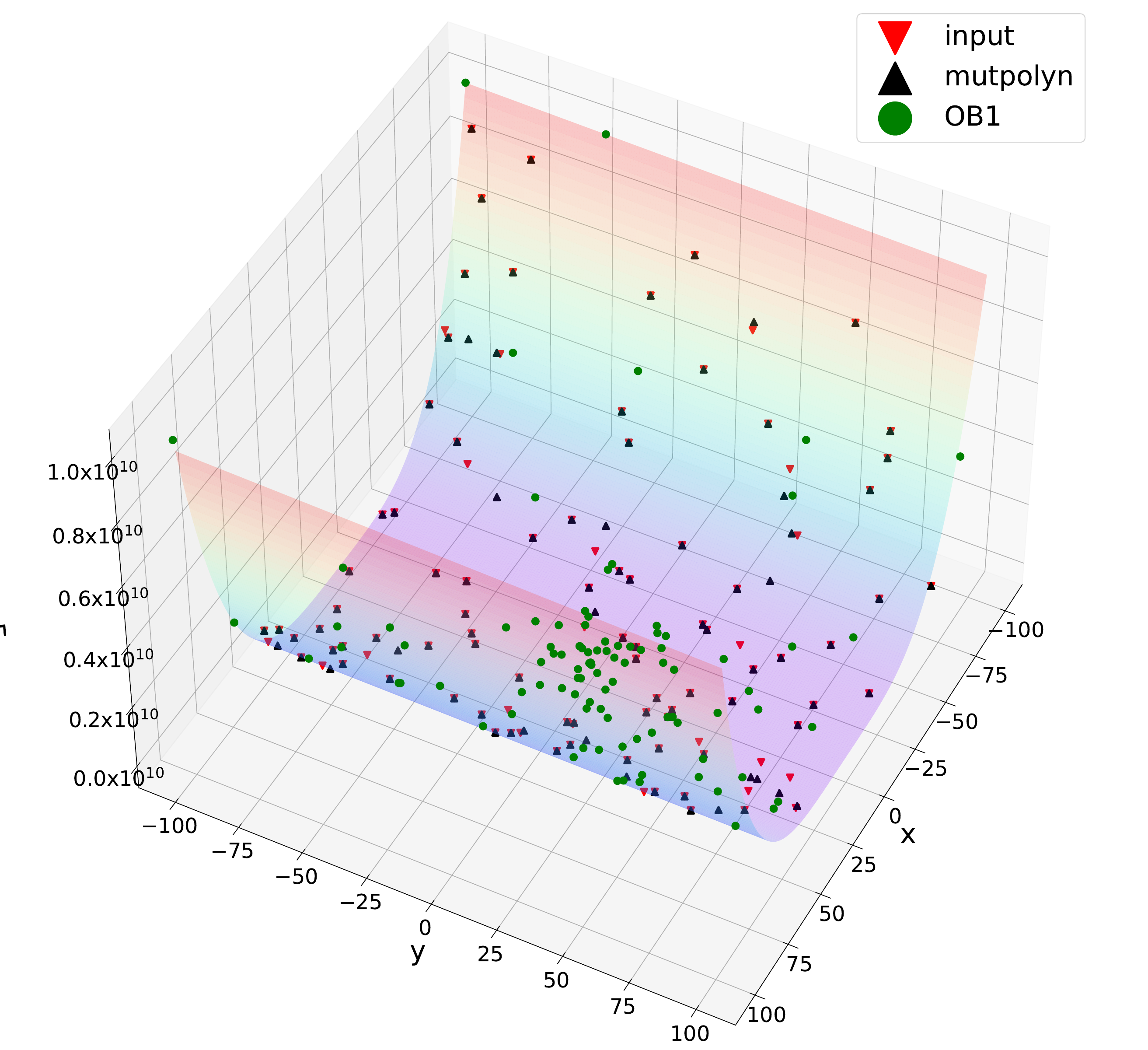}}\\
  \subfloat[]{\includegraphics[width=2.1in]{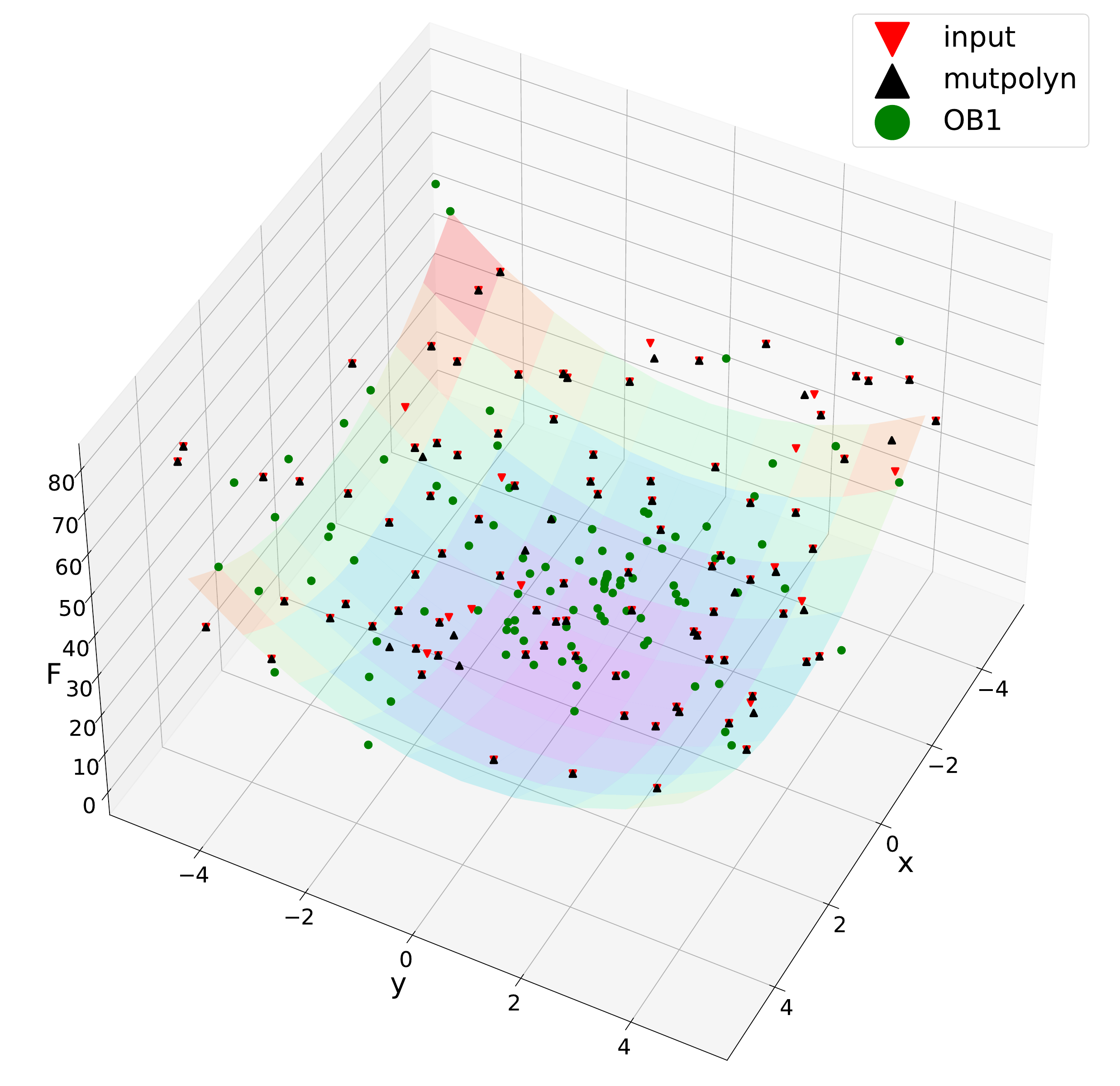}}
 \subfloat[]{\includegraphics[width=2.1in]{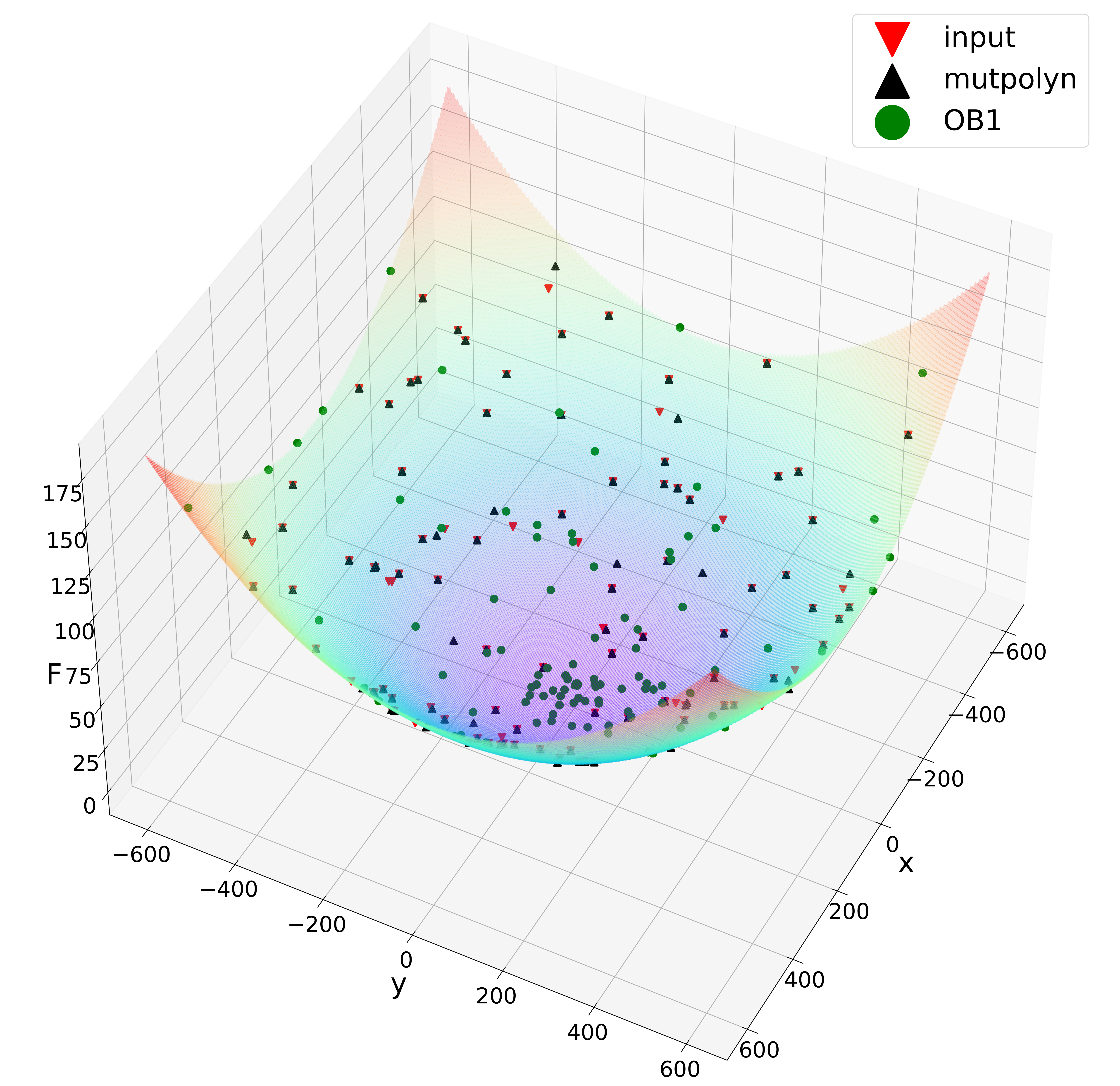}}
 \subfloat[]{\includegraphics[width=2.1in]{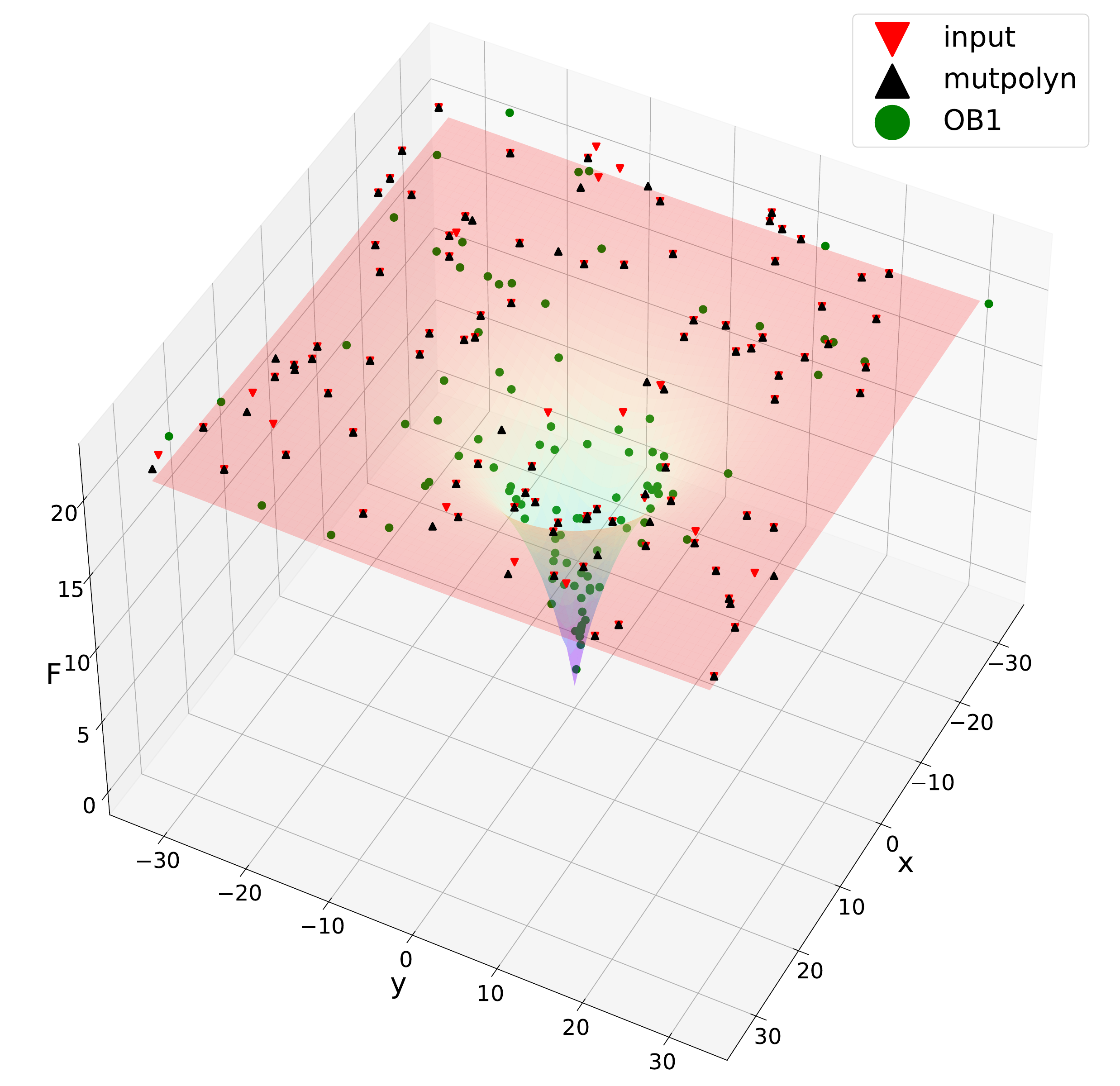}}
\caption{Mutation strategy learned by B2Opt for (a) F4, (b) F5, (c) F6, (d) F7, (e) F8, and (f) F9.}
\label{fig:3a}
\end{figure*}

\subsection{Visualization Analysis}
The tested model is \textit{5 OBs with WS} trained on F1-F3 with $d=100$. The population size is set to 100.

\textbf{Visual Analysis of SAC}
The crossover strategies learned by the five SAC are shown in Fig. \ref{fig:crossover visualization}. For the presentation, we select individuals with 1st, 50th, and 100th fitness rankings. The horizontal axis represents the fitness ranking of individuals, and the vertical axis represents the attention (weight when performing crossover) on these individuals. OB1 tends to crossover with lower-ranked individuals, showing a preference for exploration. From OB1 to OB5, the bias of SAC gradually changes from exploration to exploitation.

\textbf{Visual Analysis of FM}
We visualize the FM of B2Opt to explore its behavior as Fig. \ref{fig:3a}. As a reference, we use polynomial mutation in genetic algorithms. Given the input population (input), the mutated population (OB1) is obtained through OB1; the new population (mutpolyn) is obtained by performing polynomial mutation on the input population. We visualize F4-F9 and observe the following phenomena:

1) The population generated by performing polynomial mutation is more evenly distributed on the landscape. However, most of the solutions produced by FM in B2Opt are concentrated in "areas with greater potential", which are closer to the optimal solution. Moreover, the population distribution generated by our scheme also takes diversity into account. The non-optimal solution area is also more comprehensive than that of polynomial mutation, which is more conducive to jumping out of the local solution.

2) The population produced by performing polynomial mutation moves slightly compared to the original population. However, FM can guide the input population to make big moves toward the optimal solution, significantly accelerating the algorithm's convergence.

This shows that B2Opt can use the information of the objective function to design the mutation strategy automatically, making it more applicable to the target optimization task, which is consistent with our motivation.

\section{Conclusions}
The better performance than that of EA baselines, Bayesian optimization, and the L2O-based method demonstrates the effectiveness of B2Opt. Moreover, B2Opt can be well adapted to unseen BBO. Meanwhile, we experimentally demonstrate that the proposed three modules have positive effects. However, B2Opt still has room for improvement.
%1) Our scheme is not limited to BBO. Similar to the LSTM architecture, our scheme can directly optimize differentiable functions. However, the architecture of B2Opt does not directly involve the gradient information of the optimization target, which makes B2Opt inferior to existing L2O schemes. In future work, we will design a new module that embeds the gradient information of the optimization target;

1) In the loss function, we do not effectively consider the diversity of the population, and the population can be regularized in the future;

2) The training set seriously affects the performance of B2Opt. If the similarity between the training set and the optimization objective is low, it will cause the performance of B2Opt to degrade drastically. Building the dataset as relevant to the target as possible is essential.

\ifCLASSOPTIONcaptionsoff
  \newpage
\fi
\bibliographystyle{IEEEtran}
\bibliography{IEEEabrv,refs}

\end{document}